\documentclass[smallextended]{svjour3} 
\newif\ifdraft

\usepackage{amssymb}
\usepackage{amsmath}
\usepackage{array}
\usepackage[english]{babel}
\usepackage{color}
\usepackage{colortbl}
\usepackage{dcolumn}
\usepackage{graphicx}
\usepackage[utf8]{inputenc}
\usepackage{latexsym}
\usepackage{multicol}
\usepackage{multirow}
\usepackage[sort&compress]{natbib}
\usepackage{pifont}
\usepackage{rotating}
\usepackage{soul}
\usepackage{url}
\usepackage[usenames]{xcolor}
\usepackage{booktabs} 

\usepackage{algorithm}
\usepackage{algpseudocode}
\newcommand\commentfont{\fontfamily{pag}\scriptsize\mdseries\selectfont}

\colorlet{verylightgray}{lightgray!40}
\algtext*{EndFor}
\algtext*{EndIf}

\newcommand{\alexcomment}[1]{\ifdraft{\leavevmode\color{violet}{[AM]: {#1}}}\else{\vspace{0ex}}\fi}

\newcommand{\figext}[0]{\ifdraft png \else pdf \fi}

\newcommand{\blue}[1]{\ifdraft{\leavevmode\color{blue}{#1}}\else{\leavevmode\color{black}{#1}}\fi}

\DeclareMathOperator*{\argmin}{argmin}

\newcolumntype{M}[1]{>{\centering\arraybackslash}p{#1}}
\newcolumntype{L}{>{\centering\arraybackslash}m{12cm}}

\newif\ifplots
\plotstrue

\usepackage[normalem]{ulem}
\usepackage{nicefrac}

\newcommand{\killpunct}[1]{}

\usepackage{MnSymbol}

\DeclareMathOperator{\EX}{\mathbb{E}}

\newcommand{\gray}{\cellcolor[gray]{0.8}}

\begin{document}

\title{Binary Quantification and Dataset Shift: \\ An Experimental
Investigation}



\titlerunning{Binary Quantification and Dataset
Shift} 

\author{Pablo González 
\and Alejandro Moreo 
\and Fabrizio Sebastiani 
}

\authorrunning{P.\ González, A.\ Moreo, F.\
Sebastiani} 

\institute{Pablo González \at Artificial Intelligence Center,
University of Oviedo \\ 33204 Gijón, Spain \\
\email{gonzalezgpablo@uniovi.es} \and Alejandro Moreo and Fabrizio
Sebastiani \at Istituto di Scienza e Tecnologie dell'Informazione,
Consiglio Nazionale delle Ricerche \\ 56124 Pisa, Italy \\
\email{alejandro.moreo@isti.cnr.it, fabrizio.sebastiani@isti.cnr.it} }

\date{Received: October 2023 / Accepted: date}

\maketitle

\begin{abstract}
 \noindent Quantification is the supervised learning task that
 consists of training predictors of the class prevalence values of
 sets of unlabelled data, and is of special interest when the
 labelled data on which the predictor has been trained and the
 unlabelled data are not IID, i.e., suffer from \emph{dataset
 shift}. To date, quantification methods have mostly been tested only
 on a special case of dataset shift, i.e., \emph{prior probability
 shift}; the relationship between quantification and other types of
 dataset shift remains, by and large, unexplored. In this work we
 carry out an experimental analysis of how current quantification
 algorithms behave under different types of dataset shift, in order to 
 identify limitations of current approaches and hopefully pave the way 
 for the development of more broadly applicable methods. We do
 this by proposing a fine-grained taxonomy of types of dataset shift,
 by establishing protocols for the generation of datasets affected by
 these types of shift, and by testing existing quantification methods
 on the datasets thus generated. One finding that results from this
 investigation is that many existing quantification methods that had been found robust to prior probability shift are not necessarily
 robust to other types of dataset shift. A second finding is that no
 existing quantification method seems to be robust enough to dealing
 with all the types of dataset shift we simulate in our experiments.
 The code needed to reproduce all our experiments is publicly
 available at \url{https://github.com/pglez82/quant_datasetshift}.

\end{abstract}




\section{Introduction}
\label{sec:intro}


\noindent \emph{Quantification} (variously called \emph{learning to
quantify}, or \emph{class prior estimation}, or \emph{class
distribution estimation} -- see \citep{Esuli:2023os, Gonzalez:2017it}
for overviews) is a supervised learning task concerned with estimating
the \emph{prevalence values} (or \emph{relative frequencies}, or
\emph{prior probabilities}) of the classes in a sample of unlabelled
datapoints, using a predictive model (the \emph{quantifier}) trained
on labelled datapoints.

A straightforward solution to the quantification problem can be
obtained by (i) using a classifier to issue label predictions for the
unlabelled datapoints in the sample, (ii) counting how many datapoints
have been attributed to each class, and (iii) reporting the relative 
frequencies.
This method is typically known as \emph{Classify and Count} (CC). 
However, unless the classifier is a perfect one, CC is known to
deliver suboptimal solutions~\citep{Forman:2005fk}. One reason (but
not the only one) is that CC tends to inherit the bias of the
classifier; for example, in binary quantification problems (i.e., when
there are only two mutually exclusive classes), if the classifier has
a tendency to produce more (resp., fewer) false positives than false
negatives, CC tends to overestimate (resp., underestimate) the
prevalence of the positive class.

Since the term ``quantification'' was coined by~\citet{Forman:2005fk},
quantification has come to be recognised as a task in its own right
and is, by now, no longer considered as a mere by-product of
classification.
Quantification finds applications in many areas whose primary focus is
the analysis of data at the \emph{aggregate} level (rather at the
level of the individual datapoint), such as market
research~\citep{Esuli:2010kx}, the social
sciences~\citep{Hopkins:2010fk}, ecological
modelling~\citep{Beijbom:2015yg}, and
epidemiology~\citep{King:2008fk}, among many others.

A common trait of all these applications is that all of them emerge
from the need to monitor evolving class distributions, i.e.,
situations in which the class distribution of the unlabelled data may
differ from the one of the training data. In other words, these
situations are characterised by a type of \emph{dataset
shift}~\citep{Moreno-Torres:2012ay, Quinonero:2009kl}, i.e., the
phenomenon according to which, in a supervised learning context, the
training data and the unlabelled data are not IID.
Dataset shift comes in different flavours; the ones that have mostly
been discussed in the literature are (i) \emph{prior probability
shift}, which has to do with changes in the class prevalence values;
(ii) \emph{covariate shift}, which concerns changes in the
distribution of the covariates (i.e., features);
and (iii) \emph{concept shift}, which has to do with changes in the
functional relationship between covariates and classes. We provide
more formal definitions of dataset shift and its subtypes in the
sections to come.

Since quantification aims at estimating class prevalence, most
experimental evaluations of quantification systems (see, e.g.,
\citep{Barranquero:2015fr, Bella:2010kx, Esuli:2018rm, Forman:2008kx,
Hassan:2020kq, Milli:2013fk, Moreo:2022bf, Perez-Gallego:2019vl,
Schumacher:2021ty}) have focused on situations characterised by prior
probability shift,
while the other two types of shift mentioned above have not received
comparable attention. A question then naturally arises: \emph{How do
existing quantification methods fare when confronted with types of
dataset shift other than prior probability shift}?

This paper offers a systematic exploration of the performance of
existing quantification methods under different types of dataset
shift. To this aim we first propose a fine-grained taxonomy of dataset
shift types; in particular, we pay special attention to the case of
covariate shift, and identify variants of it (mostly having to do with
additional changes in the priors) that we contend to be of special
relevance in quantification endeavours, and that are understudied. We
then follow an empirical approach, devising specific experimental
protocols for simulating all the types of dataset shift that we have
identified, at various degrees of intensity and in a tightly
controlled manner. Using the experimental setups generated by means of
these protocols, we then test a number of existing quantification
methods; here,
the ultimate goal we pursue is to better understand the relative
merits and limitations of existing quantification algorithms, to
understand the conditions under which they tend to perform well, and
to identify the situations in which they instead tend to generate
unreliable predictions.

The rest of this paper is organised as follows. In
Section~\ref{sec:relwork}, we discuss previous work on establishing
protocols to recreate different types of dataset shift, with special
attention to work done in the quantification arena, and the (still
scarce) work aimed at drawing connections between quantification and
different types of dataset shift. In Section~\ref{sec:preliminaries},
we illustrate our notation and provide definitions of relevant
concepts and of the quantification methods we use in this
study. Section~\ref{sec:shift} goes on by introducing formal
definitions of the types of shift we investigate.
Section~\ref{sec:experiments} illustrates the experimental protocols
we propose for simulating the above types of shift, and discusses the
results we have obtained by generating datasets via these protocols
and using them for testing quantification systems.
Section~\ref{sec:conclusions} wraps up, summarising our main findings
and also pointing to interesting directions for future work.


\section{Related Work}
\label{sec:relwork}


\noindent Since quantification targets the estimation of class
frequencies, it is fairly natural that prior probability shift has
been, in the related literature, the dominant type of dataset shift on
which the robustness of quantification methods has been tested.
Indeed, when \citet{Forman:2005fk} first proposed (along with novel
quantification methods) to consider quantification as a task in its
own right (and proposed ``quantification'' as the name for this task),
he also proposed an experimental protocol for testing quantification
systems. This protocol consisted of generating a number of
test samples, to be used for evaluating a quantification method,
characterised by prior probability shift.
Given a dataset consisting of a set $L$ of labelled datapoints and a
set $U$ of unlabelled datapoints (both with binary labels), the
protocol consists of drawing from $U$ a number of test samples each
characterised by a prevalence value (of the ``positive class'') lying
on a predefined grid (say, $G=[0.00, 0.05, \ldots, 0.95, 1.00]$). This
protocol has come to be known as the ``artificial prevalence
protocol'' (APP), and has since
been at the heart of
most empirical evaluations conducted in the quantification literature;
see, e.g.,~\citep{Bella:2010kx, Barranquero:2015fr, Schumacher:2021ty,
Moreo:2021bs, Moreo:2022bf}.\footnote{Although the protocol was
originally proposed for binary quantification problems only, an
extension to the multiclass regime based on so-called \emph{Kraemer
sampling} was later proposed by \cite{Esuli:2022wf}.} Actually, the
protocol proposed by \citet{Forman:2005fk} also simulates different
prevalence values in the training set, drawing from $L$ a number of
training samples characterised by prevalence values lying on grid
$G$. In such a way, by systematically varying both the training
prevalence \textit{and} the test prevalence of the positive class
across the entire grid, one could subject a quantification method to
the widest possible range of scenarios characterised by prior
probability shift. Some empirical evaluations conducted nowadays only
extract test samples from $U$, while others extract training samples
from $L$ \emph{and} test samples from $U$.

The APP has sometimes been criticised (see e.g.,~\citep{Esuli:2015gh,
Hassan:2021bl}) for generating training-test sample pairs exhibiting
``unrealistic'' or ``implausible'' class prevalence values and/or
\alexcomment{and}
degrees of prior probability shift. For instance, \citet{Esuli:2015gh}
and \citet{Gonzalez:2019fh} indeed renounce to using the APP in favour
of using datasets containing a large amount of timestamped test
datapoints, which allows splitting the test data into sizeable enough,
temporally coherent chunks, in which the class prevalence values
naturally fluctuate over time. However, this practice is rarely used
in the literature, since it has to overcome at least three important
obstacles: (i) the amount of test samples thus available is often too
limited to allow statistically significant conclusions, (ii) datasets
with the above characteristics are rare (and expensive to create, if
not available), and (iii) the degree of shift which the quantifiers
must confront is (as in~\citep{Esuli:2015gh}) sometimes limited.

Conversely, the other two types of shift that we have mentioned above
(covariate shift and concept shift) have received essentially no
attention in the quantification literature. An exception to this
includes the theoretical analysis performed in \citep{Tasche:2022fu,
Tasche:2023um},
and the work on classifier calibration of \cite{Card:2018pb}, both of
them having to do with covariate shift. More in general, we are
unaware of the existence of specific evaluation protocols for
quantification, or quantification methods, that explicitly address
covariate shift or concept shift.

Some discussion of protocols for simulating different kinds of prior
probability shift can be found in the work of \citet{Lipton:2018fj},
who propose protocols for generating prior probability shift in
multiclass datasets. They propose protocols for addressing ``knock-out
shift'', which they define as the shift generated by subsampling a
specific class out of the $n$ classes; ``tweak-one shift'', that
generates samples in which a specific class out of the $n$ classes has
a predefined prevalence value while the rest of the probability mass
is evenly distributed across the remaining classes; and ``Dirichlet
shift'', in which a distribution $P(Y)$ across the classes is picked
from a Dirichlet distribution with concentration parameter $\alpha$,
after which samples are drawn according to $P(Y)$. Other works
\citep{Azizzadenesheli:2019qf, Rabanser:2019ba, Alexandari:2020dn}
have come to subsequently adopt these protocols. We do not explore
``knock-out shift'' nor ``tweak-one shift'' since these sample
generation protocols are only meaningful in the multiclass regime, and
since we here address the binary case only. The protocol we end up
adopting (the APP) is similar in spirit to the ``Dirichlet shift''
protocol (i.e., both are designed to cover the entire spectrum of
legitimate prevalence values), although the APP allows for a tighter
control on the test prevalence values being generated.

Using image datasets for their experiments, \cite{Rabanser:2019ba}
bring into play (and define protocols for) other types of shift having
to do with covariate shift, such as ``adversarial shift'', in which a
fraction of the unlabelled samples are adversarial samples (i.e.,
images that have been manipulated with the aim of confounding a neural
model, by means of modifications that are imperceptible to the human
eye); ``image shift'', in which the unlabelled images result from the
application of a series of random transformations (rotation,
translation, zoom-in); ``Gaussian noise shift'', in which Gaussian
noise affects a fraction of the unlabelled images; and combinations of
all these. We do not explore these types of shift since they are
specific to the world of images and computer vision.

Dataset shift has been widely studied in the field of classification
in order to support the development of models robust to the presence
of shift. In the machine learning literature this problem is also
known as \emph{domain adaptation}. For instance, the combination of
covariate shift and prior probability shift has recently been studied
by \cite{Chen:2022yb}, who focus on detecting the presence of shift in
the data and on predicting classifier performance on non-IID (a.k.a.\
``out-of-distribution'') unlabelled data. This and other similar works
are mostly concerned with improving the performance of a classifier on
non-IID unlabelled data (a concern that goes back at least
to~\citep{Saerens:2002uq, Vucetic:2001fk}, and that has given rise to
works such as \citep{Alaiz-Rodriguez:2011fk, Bickel:2009rg,
Chan2006}); in these works, estimating class prevalence in non-IID
unlabelled data is merely an intermediate step for calculating the
class weights needed for adapting the classifier to these data, and
not a primary concern in itself.

As a final note, we should mention that, despite several efforts for
unifying the terminology related to dataset shift (see
\citep{Moreno-Torres:2012ay} for an example), this terminology is
still somewhat confusing. For example, \emph{prior probability shift}
\citep{Storkey:2009lp} is sometimes called ``distribution drift''
\citep{Moreo:2022bf}, ``class-distribution shift''
\citep{Beijbom:2015yg}, ``class-prior change''
\citep{du-Plessis:2012nr,Iyer:2014ul}, ``global drift''
\cite{Hofer:2012uq}, ``target shift''~\citep{Zhang:2013iw,
Nguyen:2015kk}, ``label shift''~\citep{Lipton:2018fj,
Azizzadenesheli:2019qf, Rabanser:2019ba, Alexandari:2020dn}, or
``prior shift'' \citep{Sipka:2022aq}. The terms ``shift'' and
``drift'' are often used interchangeably (in this paper we will stick
to the former), although some authors (e.g., \cite{Souza:2020il})
establish a difference between ``concept shift'' and ``concept
drift''; in Section~\ref{sec:conceptshift} we will precisely define
what we mean by concept shift.
Note also that, until recently, most works in the quantification
literature hardly even mentioned (any type of) ``shift'' or ``drift''
(despite using an experimental protocol that recreated prior
probability shift), certainly due to the fact that the awareness of
dataset shift and the problems it entails has become widespread only
in recent years.

\section{Preliminaries}
\label{sec:preliminaries}


\subsection{Notation and Definitions}
\label{sec:notation}

\noindent In this paper we restrict our attention to the case of
binary quantification, and adopt the following notation. By
$\mathbf{x}$ we indicate a datapoint drawn from a domain
$\mathcal{X}$. By $y$ we indicate a class drawn from a set
$\mathcal{Y}=\{0,1\}$, which we call the \emph{classification scheme}
(or \emph{codeframe}), and by $\overline{y}$ we indicate the
complement of $y$ in $\mathcal{Y}$. Without loss of generality, we
assume 0 to represent the ``negative'' class and 1 to represent the
``positive'' class.
By $L$ we denote a collection of $k$ labelled datapoints
$\{(\mathbf{x}_i, y_i)\}_{i=1}^k$, where $\mathbf{x}_i\in\mathcal{X}$
is a datapoint and $y_i\in\mathcal{Y}$ is a class label, that we use
for training purposes. By $U$ we instead denote a collection
$\{(\mathbf{x}'_i, y'_i)\}_{i=1}^{k'}$ of $k'$ unlabelled datapoints,
i.e., datapoints $\mathbf{x}'_i$ whose label $y'_i$ is unknown, that
we typically use for testing purposes. We hereafter refer to $L$ and
$U$ as ``the training set'' and ``the test set'', respectively.

We use symbol $\sigma$ to denote a \emph{sample}, i.e., a non-empty
set of (labelled or unlabelled) datapoints from $\mathcal{X}$. We use
$p_{\sigma}(y)$ to denote the (true) prevalence of class $y$ in sample
$\sigma$ (i.e., the fraction of items in $\sigma$ that belong to $y$),
and we use $\hat{p}_{\sigma}^{q}(y)$ to denote the estimate of
$p_{\sigma}(y)$ as computed by a quantification method $q$; note that
$p_{\sigma}(y)$ is just a shorthand of
$P(Y=y \ | \ \mathbf{x}\in\sigma)$, where $P$ indicates probability
and $Y$ is a random variable that ranges on $\mathcal{Y}$. Since in
the binary case it holds that
$p_{\sigma}(y)=1-p_{\sigma}(\overline{y})$, binary quantification
reduces to estimating the prevalence of the positive class only.
Throughout this paper we will simply write $p_{\sigma}$ instead of
$p_{\sigma}(1)$, i.e., as a shortcut for the true prevalence of the
positive class in sample $\sigma$; similarly, we will shorten
$\hat{p}_{\sigma}(1)$ as $\hat{p}_{\sigma}$.

We define a \emph{binary quantifier} as a function
$q : 2^\mathcal{X} \rightarrow [0,1]$, i.e., one that acts as a
predictor of the prevalence $p_\sigma$ of the positive class in sample
$\sigma$.
Quantifiers are generated by means of an inductive learning algorithm
trained on $L$.
We take a (binary) \emph{hard classifier} to be a function
$h: \mathcal{X} \rightarrow \mathcal{Y}$, i.e., a predictor of the
class label of a datapoint $\mathbf{x}\in\mathcal{X}$ which returns
$1$ if
$h$ predicts $\mathbf{x}$ to belong to the positive class and $0$
otherwise. 
Classifier $h$ is trained by means of an inductive learning algorithm
that uses a set $L$ of labelled datapoints, and usually returns crisp
decisions by thresholding the output of an underlying real-valued
decision function $f$ whose internal parameters have been tuned to fit
the training data.
%
Likewise, we take a (binary) \emph{soft classifier} to be a function
$s : \mathcal{X} \rightarrow [0,1]$, i.e., a function mapping a
datapoint $\mathbf{x}$ into a \emph{posterior probability}
$s(\mathbf{x}) \equiv P(Y=1|X=\mathbf{x})$
and represents the probability that $s$ subjectively attributes to the
fact that $\mathbf{x}$ belongs to the positive class.
Classifier $s$ is either trained on $L$ by a probabilistic inductive
algorithm, or obtained by \emph{calibrating} a (possibly
non-probabilistic) classifier $s'$ also trained on $L$.\footnote{A
binary soft classifier $s$ is said to be \emph{well calibrated}
\citep{Flach:2017zp} for a given sample $\sigma$ if, for every
$\alpha\in[0,1]$, it holds that
\begin{align}
 \label{eq:calibration}
 \frac{|\{(\mathbf{x},y)\in \sigma \mid
 s(\mathbf{x})=\alpha, y=1\}|}{|\{(\mathbf{x},y)\in
 \sigma \mid s(\mathbf{x})=\alpha\}|}=\alpha
\end{align}
\noindent Note that calibration is defined with respect to a sample
$\sigma$, which means that a classifier cannot, in general, be well
calibrated for two different samples (e.g., for $L$ and $U$) that are
affected by prior probability shift.}


We take an \emph{evaluation measure} for binary quantification to be a
real-valued function $D: [0,1]\times [0,1] \rightarrow \mathbb{R}$
which measures the amount of discrepancy between the true distribution
and the predicted distribution of $\mathcal{Y}$ in $\sigma$; higher
values of $D$ represent higher discrepancy, and the distributions are
represented (since we are in the binary case) by the prevalence values
of the positive class. In the quantification literature, these
measures are typically \emph{divergences}, i.e., functions
that, given two distributions $p'$, $p''$, satisfy (i)
$D(p',p'')\geq 0$, and (ii) $D(p',p'')=0 $ if and only if $p'=p''$. By
$D(p_\sigma, \hat{p}_\sigma^q)$
%
%
%
%
we thus denote the divergence between the true class distribution in
sample $\sigma$ and the estimate of this distribution returned by
binary quantifier $q$.




\subsection{The IID Assumption, Dataset Shift, and Quantification}
\label{sec:IID}

\noindent One of the main reasons why we study quantification is the
fact that most scenarios in which estimating class prevalence values
via supervised learning is of interest, \emph{violate the IID
assumption}, i.e., the fundamental assumption (that most machine
learning endeavours are based on) according to which the labelled
datapoints used for training and the unlabelled datapoints we want to
issue predictions for, are assumed to be drawn independently and
identically from the same (unknown) distribution.\footnote{For
example, we might be interested in monitoring through time the degree
of support for a certain politician by estimating the prevalence
values of classes ``Positive'' and ``Negative'' in tweets that express
opinions about this politician (this is an instance of \emph{sentiment
quantification}~\citep{Moreo:2022bf}). The very fact that we want to
monitor these prevalence values through time is an implicit assumption
that these prevalence values may vary, i.e., may take values different
from the prevalence values that these classes had in the training
data. In other words, it is an implicit assumption that we may be in
the presence of some form of dataset shift.} If the IID assumption
were not violated, the supervised class prevalence estimation problem
would admit a trivial solution, consisting of returning, as the
estimated prevalence $\hat{p}_{\sigma}^{q}$ for \emph{any} sample
$\sigma$ of unlabelled datapoints, the true prevalence $p_{L}$ that
characterises the training set, since both $L$ and $\sigma$ would be
expected to display the same prevalence values. This ``method'' is
called, in the quantification literature, the \emph{maximum likelihood
prevalence estimator} (MLPE), and is considered a trivial baseline
that any genuine quantification system is expected to beat in
situations characterised by dataset shift.

We will thus assume the existence of two unknown joint probability
distributions $P_{L}(X,Y)$ and $P_{U}(X,Y)$ such that
$P_{L}(X,Y)\neq P_{U}(X,Y)$ (the \emph{dataset shift assumption}). The
ways in which the training distribution and the test distribution may
differ, and the effect these differences can have on the performance
of quantification systems, will be the main subject of the following
sections.


\subsection{Quantification Methods}
\label{sec:methods}

\noindent The six quantification methods that we use in the
experiments of Section~\ref{sec:experiments} are the following.

\emph{Classify and Count} (CC), already hinted at in the introduction,
is the na\"ive quantification method, and the one that is used as a
baseline that all genuine quantification methods are supposed to
beat. Given a hard classifier $h$ and a sample $\sigma$, CC is
formally defined as
\begin{align}
 \begin{split}
 \label{eq:CC}
 \hat{p}_{\sigma}^{\mathrm{CC}}=\frac{1}{|\sigma|}\sum_{\mathbf{x}\in
 \sigma}h(\mathbf{x})
 \end{split}
\end{align}
\noindent In other words, the prevalence of the positive class is
estimated by classifying all the unlabelled datapoints, counting the
number of datapoints that have been assigned to the positive class,
and dividing the result by the total number of datapoints in the
sample.

The \emph{Adjusted Classify and Count} (ACC) method
(see~\citep{Forman:2008kx})
attempts to correct the estimates returned by CC by relying on the law
of total probability, according to which, for any
$\mathbf{x}\in\mathcal{X}$, it holds that
\begin{align}
 \label{eq:ACC} 
 P(h(\mathbf{x})=1) & = P(h(\mathbf{x})=1|Y=1)\cdot p 
 + P(h(\mathbf{x})=1|Y=0)\cdot (1-p)
\end{align}
\noindent which can be more conveniently rewritten as
\begin{align}
 \begin{split}
 \label{eq:acc2}
 \hat{p}_{\sigma}^{\mathrm{CC}} \ & = \ \operatorname{tpr}_{h}\cdot
 p_{\sigma} + \operatorname{fpr}_{h}\cdot (1-p_{\sigma})
 \end{split}
\end{align}
\noindent where $\operatorname{tpr}_{h}$ and $\operatorname{fpr}_{h}$
are the true positive rate and the false positive rate, respectively,
that $h$ has on samples of unseen datapoints. From
Equation~\ref{eq:acc2} we can obtain
\begin{align}
 \label{eq:acc3}
 p_{\sigma}=
 \frac{\hat{p}_{\sigma}^{\operatorname{CC}}
 -\operatorname{fpr}_{h}}{\operatorname{tpr}_{h}-\operatorname{fpr}_{h}}
\end{align}
\noindent The values of $\operatorname{tpr}_{h}$ and
$\operatorname{fpr}_{h}$ are unknown, but their estimates
$\hat{\operatorname{tpr}}_{h}$ and $\hat{\operatorname{fpr}}_{h}$ can
be obtained by performing $k$-fold cross-validation on the training
set $L$, or by using a held-out validation set. The ACC method thus
consists of estimating $p_{\sigma}$ by plugging the estimates of tpr
and fpr into Equation~\ref{eq:acc3}, to obtain
\begin{align}
 \label{eq:acc4}
 \hat{p}_{\sigma}^{\operatorname{ACC}}=
 \frac{\hat{p}_{\sigma}^{\operatorname{CC}}
 -\hat{\operatorname{fpr}}_{h}}{\hat{\operatorname{tpr}}_{h}-\hat{\operatorname{fpr}}_{h}}
\end{align}
\noindent While CC and ACC rely on the crisp counts returned by a hard
classifier $h$, it is possible to define variants of them that use
instead the \emph{expected} counts computed from the posterior
probabilities returned by a calibrated probabilistic classifier $s$
\citep{Bella:2010kx}. This is the core idea behind \emph{Probabilistic
Classify and Count} (PCC) and \emph{Probabilistic Adjusted Classify
and Count} (PACC).
PCC is defined as
\begin{align}
 \begin{split}\label{eq:pcc}
 \hat{p}_{\sigma}^{\mathrm{PCC}} & = \frac{1}{|\sigma|}\sum_{\mathbf{x}\in \sigma}s(\mathbf{x}) \\
 & = \frac{1}{|\sigma|}\sum_{\mathbf{x}\in \sigma}P(Y=1|\mathbf{x})
 \end{split}
\end{align}
\noindent while PACC is defined as
\begin{align}
 \label{eq:pacc}
 \hat{p}_{\sigma}^{\operatorname{PACC}}=
 \frac{\hat{p}_{\sigma}^{\operatorname{PCC}}
 -\hat{\operatorname{fpr}}_{s}}{\hat{\operatorname{tpr}}_{s}-\hat{\operatorname{fpr}}_{s}}
\end{align}
\noindent Equation~\ref{eq:pacc} is identical to
Equation~\ref{eq:acc4}, but for the fact that the estimate
$\hat{p}_{\sigma}^{\operatorname{CC}}$ is replaced with the estimate
$\hat{p}_{\sigma}^{\operatorname{PCC}}$, and for the fact that the
true positive rate and the false positive rate of the probabilistic
classifier $s$ (i.e., the rates computed as expectations using the
posterior probabilities) are used in place of their crisp
counterparts.

\emph{Distribution y-Similarity} (DyS)~\citep{Maletzke:2019qd} is
instead a generalisation of the HDy quantification method of
\citet{Gonzalez-Castro:2013fk}. HDy is a probabilistic binary
quantification method that views quantification as the problem of
minimising the divergence (measured in terms of the Hellinger
Distance, from which the name of the method derives) between two
distributions of posterior probabilities returned by a soft classifier
$s$, one coming from the unlabelled examples and the other coming from
a validation set. HDy looks for the mixture parameter $\alpha$ (since
we are considering a mixture of two distributions, one of examples of
the positive class and one of examples of the negative class) that
best fits the validation distribution to the unlabelled distribution,
and returns $\alpha$ as the estimated prevalence of the positive
class. Here, robustness to distribution shift is achieved by the
analysis of the distribution of the posterior probabilities in the
unlabelled set, that reveals how conditions have changed with respect
to the training data. DyS generalises HDy by viewing the divergence
function to be used as a parameter.


A further, very popular aggregative quantification method is the one
proposed by~\citet{Saerens:2002uq} and often called SLD, from the
names of its proposers. SLD was the best performer in a recent data
challenge devoted to quantification~\citep{Esuli:2022wf}, and consists
of training a (calibrated) soft classifier and then using expectation
maximisation \citep{Dempster:1977hl} (i) to tune the posterior
probabilities that the classifier returns, and (ii) to re-estimate the
prevalence of the positive class in the unlabelled set. Steps (i) and
(ii) are carried out in an iterative, mutually recursive way, until
convergence
(when the estimated prior gets fairly close to the
mean of the recalibrated posteriors).
%
%



\section{Types of Dataset Shift}
\label{sec:shift}

\noindent Any joint probability distribution $P(X,Y)$ can be
factorised, alternatively and equivalently, as:

\begin{itemize}
\item $P(X,Y)=P(X|Y)P(Y)$, in which the marginal distribution $P(Y)$
 is the distribution of the class labels, and
 the conditional distribution $P(X|Y)$ is the class-conditional
 distribution of the covariates. This factorization is convenient in
 \textit{anti-causal learning} (i.e., when predicting causes from
 effects) ~\citep{Scholkopf:2012je}, i.e., in \textit{problems of
 type} $Y\rightarrow X$ ~\citep{Fawcett:2005fk}.
\item $P(X,Y)=P(Y|X)P(X)$, in which the marginal distribution $P(X)$
 is the distribution of the covariates and the conditional
 distribution $P(Y|X)$ is the distribution of the labels conditional
 on the covariates.
 This factorization is convenient in \textit{causal learning} (i.e.,
 when predicting effects from causes)~\citep{Scholkopf:2012je}, i.e.,
 in \textit{problems of type}
 $X\rightarrow Y$~\citep{Fawcett:2005fk}.
\end{itemize}
Which of these four ingredients (i.e., $P(X)$, $P(Y)$, $P(X|Y)$,
$P(Y|X)$) change or remain the same across $L$ and $U$, gives rise to
different types of shift, as discussed in \citep{Storkey:2009lp,
Moreno-Torres:2012ay}.
In this section we turn to describing the types of shift that we
consider in this study. To this aim, also recalling that the related
terminology is sometimes confusing in this respect (as also noticed by
\citet{Moreno-Torres:2012ay}), we clearly define each type of shift
that we consider.

When training a model, using our labelled data, to issue predictions
about unlabelled data, we expect some relevant general conditions to
be invariant across the training distribution and the unlabelled
distribution, since otherwise the problem would be unlearnable. In
Table~\ref{tab:datasetshift}, we list the three main types of dataset
shift that have been discussed in the literature. For each such type,
we indicate which distributions are assumed (according to general
consensus in the field) to vary across $L$ and $U$, and which others
are assumed to remain constant. In the following sections, we will
thoroughly discuss the relationships between these three types of
shift and quantification.
\begin{table}[t!]
 \centering
 \begin{tabular}{l||c|c|c|c||c}
 & $P(X)$ & $P(Y)$ & $P(X|Y)$ & $P(Y|X)$ & Section \\ 
 \hline \hline
 Prior probability shift & & $\neq$ & = & & \S\ref{sec:priorprobabilityshift} \\
 \hline 
 Covariate shift & $\neq$ & & & = & \S\ref{sec:covariateshift} \\ 
 \hline 
 Concept shift & & & $\neq$ & $\neq$ & \S\ref{sec:conceptshift} \\
 \hline
 \end{tabular}
 \caption{Main types of dataset shift discussed in the literature.
 For the type of dataset shift on the row, symbol ``$\neq$''
 indicates that the distribution on the column is assumed to change
 across $L$ and $U$, while symbol $=$ indicates that the distribution
 is assumed to remain invariant. The last column indicates the
 section of the present paper where this type of shift is discussed
 in detail.}
 \label{tab:datasetshift}
\end{table}


It is immediate to note from Table~\ref{tab:datasetshift} that, for
any given type of shift, there are some distributions (corresponding
to the blank cells in the table -- e.g., $P(X)$ for prior probability
shift) for which it is not specified if they change or not across $L$
and $U$; indeed, concerning what happens in these cases, the
literature is often silent. In the next sections, we will try to fill
these gaps. We will identify applicatively interesting subtypes of
dataset shift based on different ways to fill the blank cells of
Table~\ref{tab:datasetshift}, and will propose experimental protocols
that recreate them in order for quantification systems to be tested
under those conditions.


%



\subsection{Prior Probability Shift}
\label{sec:priorprobabilityshift}

\begin{figure}[t!]
 \begin{center}
 \includegraphics[width=\textwidth]{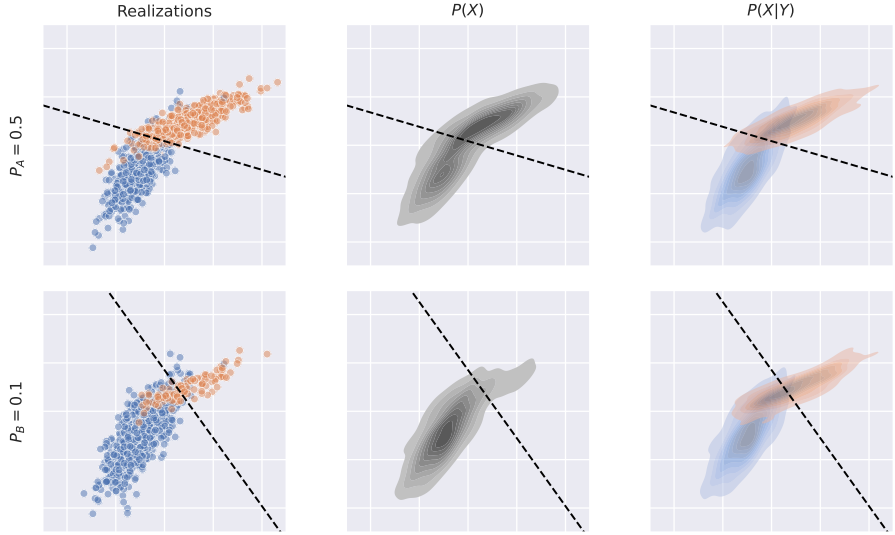}
 \end{center}
 \caption{Example of prior probability shift generated with synthetic data using a normal distribution for each class. Scenario $A$ (1st
 row): original data distribution, in which the positive class (orange)
 and the negative class (blue) have the same prevalence, i.e.,
 $p_{A}=0.5$. Scenario $B$ (2nd row): with respect to Scenario $A$ 
 there is a shift in the
 prevalence such that
 $p_{B}=0.1$. Dashed lines represent linear hypotheses learnt
 from the corresponding empirical distributions. Note that, although
 the positive class and the negative class may have not changed in meaning
 between $A$ and $B$, i.e., $P_{A}(Y|X)=P_{B}(Y|X)$, the posteriors
 we would obtain by calibrating two soft classifiers trained from the
 two empirical distributions would likely differ. Note also that
 $P_{A}(X) \neq P_{B}(X)$ (2nd column) but $P_{A}(X|Y) = P_{B}(X|Y)$
 (3rd column). 
 }
 
 \label{fig:prior1}
\end{figure}

\noindent \emph{Prior probability shift} (see Figure~\ref{fig:prior1}
for a graphical example) describes a situation in which (a) there is a
change in the distribution $P(Y)$ of the class labels (i.e.,
$P_{L}(Y) \neq P_{U}(Y)$) while (b) the class-conditional distribution
of the covariates remains constant (i.e., $P_{L}(X|Y)=P_{U}(X|Y)$).
%

In this type of shift, no further assumption is usually made as to
whether the distribution $P(X)$ of the covariates and the conditional
distribution $P(Y|X)$ change or not across $L$ and
$U$. Notwithstanding this, it is reasonable to think that
the change in $P(Y)$ indeed causes a
variation in $P(X)$, i.e., that $P_{L}(X)\not =P_{U}(X)$;
if this were not the case, the class-conditional distributions
$P(X|Y=1)$ and $P(X|Y=0)$ would be indistinguishable, i.e., the
problem would not be learnable.
We will thus assume that prior probability shift does indeed imply a
change in $P(X)$ across $L$ and $U$. The following is an example of
this scenario.

\begin{example}
 Assume our application is one of handwritten digit
 recognition. Here, the classes are all the possible types of digits
 and the covariates are features of the handwritten realizations of
 these digits. Assume our training data are (labelled) handwritten
 digits in the decimal system (digits from 0 to 9) while our
 unlabelled data are handwritten digits in the binary system (0 or
 1); assume also that all other properties of the data (e.g., authors
 of these handwritings, etc.) are the same as in the training
 data. In this scenario, it is the case that $P_{L}(Y) \neq P_{U}(Y)$
 (since, e.g., the prevalence values in $U$ of the digits from 2 to 9
 are all equal to 0, unlike in $L$), and it is the case that
 $P_{L}(X|Y)=P_{U}(X|Y)$ (since the 0's and 1's in the unlabelled
 data look the same as the 0's and 1's in the training
 data). Therefore, this is an example of prior probability
 shift. Note that it is also the case that $P_{L}(X)\not = P_{U}(X)$,
 since in $P_{U}(X)$ the values of the covariates are just those
 typical of 0's and 1's, unlike in $P_{L}(Y)$, and it is also the
 case that $P_{L}(Y|X)=P_{U}(Y|X)$, since nothing in the functional
 relationship between $X$ and $Y$ has changed.
 \qed
\end{example}
Concerning the issue of whether, in prior probability shift, the
posterior distribution $P(Y|X)$ is invariant or not across $L$ and
$U$, it seems\blue{, at first glance,} sensible to assume that
it indeed is,
i.e., $P_{L}(Y|X)=P_{U}(Y|X)$, since there is nothing in prior
probability shift that implies 
a change in the
functional relationship between $X$ and $Y$ (in the binary case: in
what being a member of the positive class or of the negative class
actually means). 
\blue{However, it turns out that a change in the priors has an impact
on the \emph{a posteriori} distribution of the response variable $Y$, i.e., that $P_{L}(Y|X) \neq P_{U}(Y|X)$.
This is indeed the reason why the posterior probabilities issued 
by a probabilistic classifier $s$ (which has been trained and calibrated for the training distribution) would need to be recalibrated for the target distribution 
before attempting to estimate $P_U(Y)$ as $\frac{1}{|U|}\sum_{\mathbf{x}\in U} s(\mathbf{x})$.
This is exactly the rationale behind the SLD method proposed by \cite{Saerens:2002uq}.
}
Following this assumption, prior probability shift is
defined as in Row 1 of Table~\ref{tab:oursetting}.

Prior probability shift is the type of shift which quantification
methods have mostly been tested on, and the invariance assumption
$P_{L}(X|Y)=P_{U}(X|Y)$ that is made in prior probability shift indeed
guarantees that a number of quantification methods work well in these
scenarios.
In order to show this, let us take ACC as an example. The correction
implemented in Equation~\ref{eq:acc4} does not attempt to counter
prior probability shift, but attempts to counter classifier bias
(indeed, note that this correction is meaningful even in the absence
of prior probability shift). This adjustment relies on
Equation~\ref{eq:acc2}, which depends on two quantities, the tpr and
the fpr of classifier $h$, that must be estimated on the training data
$L$.
Since $h(\mathbf{x})$ is the same for $L$ and $U$, the fact that
$P_{L}(X|Y)=P_{U}(X|Y)$ (which is assumed to hold under probability
shift) implies that
$\hat{\operatorname{tpr}}_{h}=\operatorname{tpr}_{h}$ and
$\hat{\operatorname{fpr}}_{h}=\operatorname{fpr}_{h}$. In other words,
under prior probability shift ACC works well, since the assumption
that the class-conditional distribution $P(X|Y)$ is invariant across
$L$ and $U$ guarantees that our estimates of tpr and fpr are good
estimates. Similar considerations apply to different quantification
methods as well.

Prior probability shift has been widely studied in the quantification
literature, both from a theoretical point of view
\citep{Tasche:2017ij, Vaz:2019eu}
and from an empirical point of view \citep{Schumacher:2021ty}. Indeed,
note that the artificial prevalence protocol (APP -- see Section~\ref{sec:relwork}), 
on which most experimentation of quantification systems has been based, 
does nothing else than generate a set of samples characterised by prior probability shift with respect
to the set from which they have been extracted; the APP recreates the
$P_{L}(Y) \neq P_{U}(Y)$ condition by subsampling \emph{one} of the
two classes, and recreates the $P_{L}(X|Y)=P_{U}(X|Y)$ condition by
performing this subsampling in a random fashion.

Most of the quantification literature is concerned with ways of
devising robust estimators of class prevalence values in the presence
of prior probability shift.
\cite{Tasche:2017ij} proves that, when $P_{L}(Y) \neq P_{U}(Y)$ and
$P_{L}(X|Y) = P_{U}(X|Y)$ (i.e., when we are in the presence of prior
probability shift)
the method ACC is \emph{Fisher-consistent}, i.e., the error of ACC
tends to zero when the size of the sample increases.
Unfortunately, in practice, the condition of an unchanging $P(X|Y)$ is
difficult to fulfil or verify.

At this point, it may be worth stressing that not every change in
$P(Y)$ can be considered an instance of prior probability
shift. Indeed, in Section~\ref{sec:covariateshift} we present
different cases of shift in the priors that are \emph{not} instances
of prior probability shift, and that we deem of particular interest
for realistic applications of quantification.
%


\subsection{Covariate Shift}
\label{sec:covariateshift}

\noindent
\emph{Covariate shift} (see Figure~\ref{fig:covariate1} for a
graphical example) describes a situation in which (a) there is a
change in the distribution $P(X)$ of the covariates (i.e.,
$P_{L}(X)\neq P_{U}(X)$), while (b) the distribution of the classes
conditional on the covariates remains constant (i.e.,
$P_{L}(Y|X) = P_{U}(Y|X)$). In this type of shift, no further
assumption is usually made as to whether the distribution $P(Y)$ of
the classes and the class-conditional distribution $P(X|Y)$ change
across $L$ and $U$.

\begin{figure}[t!]
 \centering
 \includegraphics[width=0.8\textwidth]{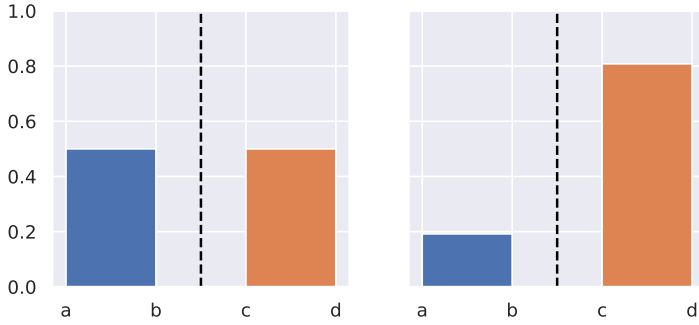}
 \caption{Possible overlap between the notions of prior probability shift and
 covariate shift, unless we assume that $P_{L}(X|Y)\neq P_{U}(X|Y)$ in
 covariate shift.}
 \label{fig:priororcovariate}
\end{figure}

In this paper, we are going to assume that
also a change in the class-conditional distribution takes place, i.e.,
$P_{L}(X|Y)\neq P_{U}(X|Y)$. The rationale of this choice is that,
without this assumption, there would be a possible overlap between the
notion of prior probability shift and the notion of 
covariate shift. To see why,
imagine a situation in which the positive and the negative examples
are numerical univariate data each following a uniform distribution
$\mathbf{U}(a,b)$ and $\mathbf{U}(c,d)$, with different parameters
$a<b<c<d$. A change in the priors (i.e., $P_{L}(Y)\neq P_{U}(Y)$)
would not cause any modification in the class-conditional distribution
(i.e., $P_{L}(X|Y)=P_{U}(X|Y)$ would hold).
Thus, by definition, this would squarely count as an example of prior
probability shift, since these are the same conditions listed in Row 1
of Table~\ref{tab:oursetting}. However, at the same time, the
distribution of the covariates has also changed (i.e.,
$P_{L}(X)\neq P_{U}(X)$), since
$P(X)=\mathbf{U}(a,b)P(Y=1)+\mathbf{U}(c,d)P(Y=0)$ and since the
priors have changed, with the posterior distribution $P(Y|X)$
remaining stable across $L$ and $U$. Thus, this would \emph{also}
count as an example of covariate shift; see
Figure~\ref{fig:priororcovariate} for a graphical explanation. For
this reason, and for the sake of clarity in the exposition, in this
work we will break the ambiguity by assuming that covariate shift
implies that $P(X|Y)$ is \textit{not} invariant across $L$ and $U$. 
As a final observation, note that the conditions of covariate shift 
are incompatible with a situation in which both 
$P(Y)$ and $P(X|Y)$ remain invariant.
The reason is that $P(X)$ is assumed to change under the covariate shift assumptions,  
but, since $P(X)=P(X|Y=1)P(Y=1)+P(X|Y=0)P(Y=0)$, 
the only way in which this condition can hold true 
comes down to assuming either a change in $P(Y)$ or in $P(X|Y)$.

We will further distinguish between two types of covariate shift,
i.e., (i) \emph{global} covariate shift, in which the changes in the
covariates occur globally, i.e., affect the entire population, and
(ii) \emph{local} covariate shift, in which the changes in the
covariates occur locally, i.e., only affect certain subregions of the
entire population. These two types of covariate shift will be the
subject of Sections~\ref{sec:globalcovariateshift}
and~\ref{sec:localcovariateshift}, respectively.

\subsubsection{Global Covariate Shift}
\label{sec:globalcovariateshift}

\noindent \emph{Global covariate shift} occurs when there is an
overall change in the representation function. We will study two
variants of it that differ in terms of whether $P(Y)$ is invariant or not across $L$ and $U$:
\emph{global pure covariate shift}, in which $P_{L}(Y)=P_{U}(Y)$, and
\emph{global mixed covariate shift}, in which $P_{L}(Y)\neq
P_{U}(Y)$ (the name ``mixed'' of course refers to the fact that there 
is a change in the distribution of the covariates \textit{and} in the 
distribution of the labels). Both scenarios are interesting to test 
quantification methods on, but the latter is probably even more 
interesting, since changes in the priors are something 
that quantification methods are expected to be robust to.


Global pure covariate shift might occur when, for example, a sensor
(in charge of generating the covariates) experiences a change (e.g., a
partial damage, or a change in the lighting conditions for a camera);
in this case, the prevalence values of the classes of interest do not change, but the measurements (covariates) might have been
affected.\footnote{\blue{This example is what \cite{kull2014patterns} called \emph{covariate observation shift}.}} 

Global mixed covariate shift might occur when, for example, a
quantifier is trained to monitor the proportion of positive opinions
on a certain politician on Twitter on a daily basis and, after the
quantifier has been deployed to production, there is a change in
Twitter's policy that allows for longer tweets.\footnote{This actually
happened in 2017, when Twitter raised the maximum allowed size of 
tweets from 140 to 280 characters.} In this case, there is a variation 
in $P(X)$, since longer tweets become more likely; there is variation in
$P(X|Y)$, since there will likely be longer positive tweets and longer
negative tweets; $P(Y|X)$ will remain constant, since a change in the
length of tweets does not make positive comments more likely or less
likely; 
and $P(Y)$ can change too (because opinions on politicians do change
in time), although not as a result of the change in tweet length.
\alexcomment{A problem of this definition is that the space is not
``absolutely continuous'', which I seem to remember is a condition for covariate shift according to Tasche.}

\begin{figure}[ht!]
 \begin{center}
 \includegraphics[width=\textwidth]{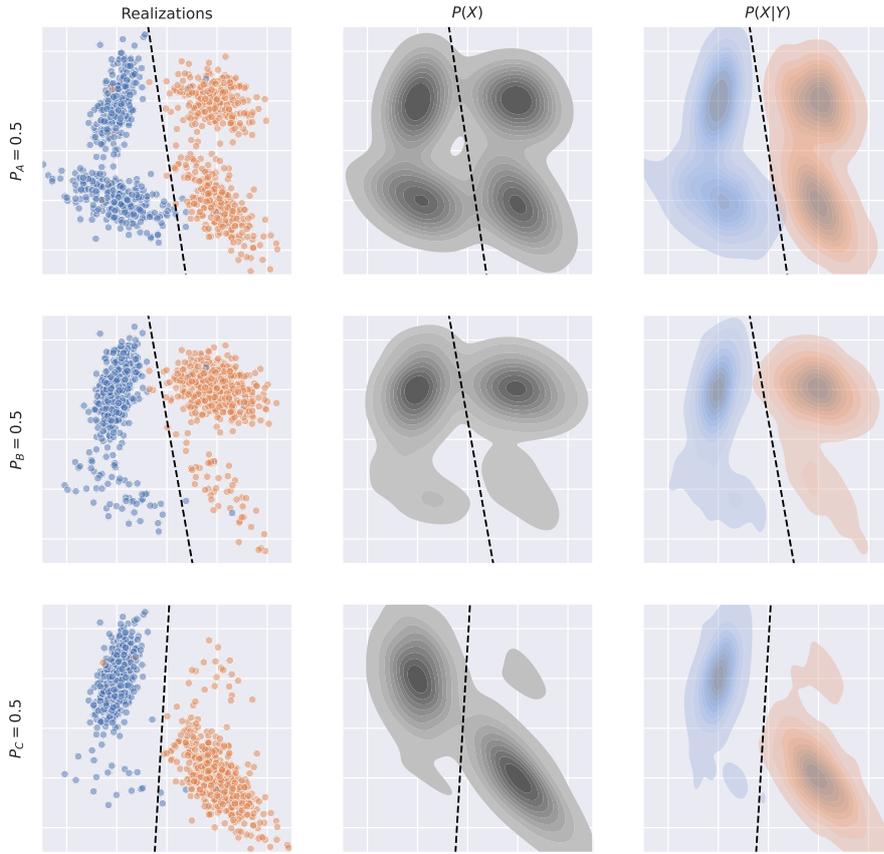}
 \end{center}
 \caption{Example of global pure covariate shift generated with
 synthetic data using a normal distribution for each
 cluster. Situation $(a)$ (1st row): original data distribution. Each
 class consists on two clusters of data (for example positive or
 negative opinions of two different categories: \textsc{Electronics}
 and \textsc{Books}). Situation $(b)$ (2nd row): there is a shift in
 the number of opinions of one category, that affects both
 classes. $P(X)$ changes (see 2nd column) but $P(Y|X)$ remains
 invariant. Situation $C$ (3rd row), $P(X)$ changes abruptly,
 affecting the posterior probabilities $s(\mathbf{x})$ 
 that a soft classifier, trained via induction on this scenario,
 would issue.}
 \label{fig:covariate1}
\end{figure}

By taking into account the underlying conditions of pure covariate
shift, it seems pretty clear that PCC (see Section~\ref{sec:methods})
would represent the best possible choice. The reason is that PCC
computes the estimate of the class prevalence values by relying on the
posterior probabilities returned by a soft classifier $s$ (see
Equation~\ref{eq:pcc}). Inasmuch as these posterior probabilities are
reliable enough (i.e., when the soft classifier is well calibrated
\citep{Card:2018pb}), the class prevalence values would be well
estimated without further manipulations (i.e., there is no need to
adjust for possible changes in the priors since, in the pure version,
we assume $P(Y)$ has not changed); see Figure~\ref{fig:covariate1},
2nd row.

However, in practice, the posterior probabilities returned by $s$
might not align well with the underlying concept of the positive class
(the soft classifier $s$ might not be well calibrated for the
unlabelled distribution). This might be due to several reasons, but a
relevant possibility is due to the inability of the learning device to
find good parameters for the classifier. This might happen whenever
the hypothesis (i.e., the soft classifier $s$) learnt by means of an
inductive learning method (e.g., logistic regression) comes from an
empirical distribution in which certain regions of the input space
were insufficiently represented during training, and have later become
more prevalent during test as a result of a change in $P(X)$; see
Figure~\ref{fig:priororcovariate}, 3rd row. This situation is
certainly problematic, and would lead to a deterioration in
performance of most aggregative quantifiers (including PCC).
%
%
An in-depth theoretical analysis of the implications of pure covariate
shift is offered by \citet{Tasche:2022fu}, in which the author
concludes that in order for PCC to prove resilient to covariate shift,
the test distribution should be \emph{absolutely continuous} with
respect to the training distribution.
This assumption also restricts the amount of divergence between
training and test samples distributions. However, these are
theoretical considerations that are hard (if not impossible) to verify
in practice.

\subsubsection{Local Covariate Shift}
\label{sec:localcovariateshift}

\begin{figure}[t!]
 \begin{center}
 \includegraphics[width=\textwidth]{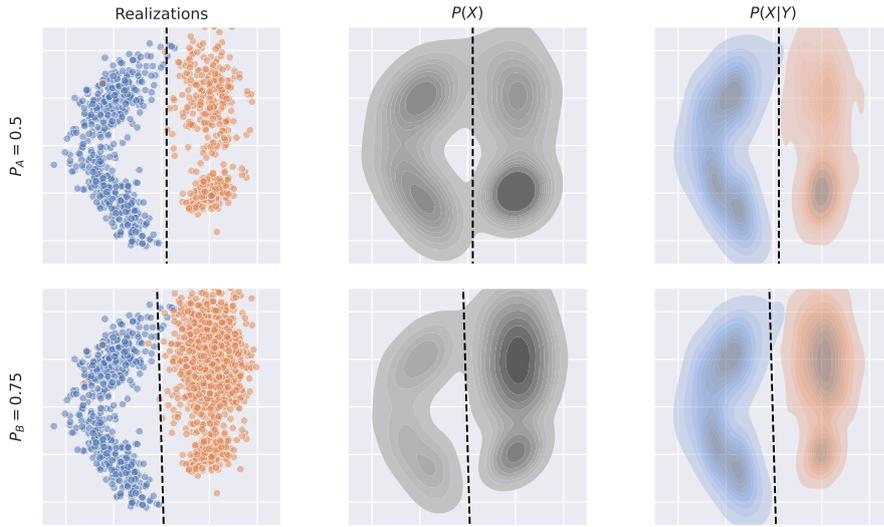}
 \end{center}
 \caption{Example of \emph{local} covariate shift generated with
 synthetic data using a normal distribution for each
 cluster. Situation (a) (1st row): original data distribution with
 two positive (orange) Gaussians $\mathcal{N}_{1}$, $\mathcal{N}_{2}$
 and two negative (blue) Gaussians $\mathcal{N}_{3}$,
 $\mathcal{N}_{4}$. Situation (b) (2nd row): the prevalence of
 $\mathcal{N}_{1}$ grows.}
 \label{fig:covariatet2}
\end{figure}

\noindent Consider a binary problem in which the positive class is a
mixture of two (differently parameterised) Gaussians $\mathcal{N}_{1}$
and $\mathcal{N}_{2}$, i.e., that
$P(X|Y=1)=\alpha \mathcal{N}_{1} + (1-\alpha) \mathcal{N}_{2}$. Assume
there are analogous Gaussians $\mathcal{N}_{3}$ and $\mathcal{N}_{4}$
governing the distribution of negatives; see
Figure~\ref{fig:covariatet2}. Assume now that there is a change (say,
an increase) in the prevalence of datapoints from $\mathcal{N}_{1}$
leading to an overall change in the priors $P(Y)$. Note that this also
implies an overall change in $P(X)$. There is also a change in
$P(X|Y=1)$ (therefore, in $P(X|Y)$) since the parameter $\alpha$ of
the mixture has changed (it is now more likely to find positive
examples from $\mathcal{N}_{1}$). However, the change in the
covariates is \emph{asymmetric}, i.e., $P(X|Y=0)$ has not changed.

Situations like this naturally occur in real scenarios of interest for
quantification. For example, in ecological modelling, researchers
might be interested in estimating the prevalence of, e.g., different
species of plankton in the sea. To do so, they analyse pictures of
water samples taken by an automatic optical device, identify
individual examplars of plankton, and estimate the prevalence of the
different species via a quantifier
\citep{Gonzalez:2019fh}. However, these plankton species are typically
grouped, because of their high number, into coarse-grained
superclasses (i.e., parent nodes from a taxonomy of classes), which
means that no prevalence estimation for the subclasss is attempted. An
increase in the prevalence value of one of the (super-)classes is
often the consequence of an increase in the prevalence value of only
one of its (hidden) subclasses.
A similar example may be found in seabed cover mapping for coral reef
monitoring~\citep{Beijbom:2015yg}; here, ecologists are interested in
quantifying the presence of different species in images, often
grouping the coral species and algae species into coarser-grained
classes.

In contrast to global covariate shift, local covariate shift does not
occur due to a variation in the feature representation function (e.g.,
an alteration of the device in charge of taking measurements, which
would impact on the covariates) but due to changes in the priors of
(sub-)classes that remain hidden. The most important implication for
quantification concerns the fact that this shift would reduce to prior
probability shift if the subclasses (the original species in our
examples) were observed in place of the
superclasses.\footnote{Technically speaking, any distribution can be
expressed as a (potentially infinite) mixture of Gaussians; thus, in
theory, one could always reduce the problem to prior probability
shift. In our definition, however, we assume the existence of a
limited set of real subpopulations with unobserved labels, and not of
an infinite such set.}
We will only consider the case in which $P(Y)$ changes, since it is
hard to think of any realistic scenario for asymmetric covariate shift
in which the class prevalence values remain unaltered.
\blue{Note also that, in extreme cases, an abrupt change in $P(Y)$ can end up compromising the condition $P_L(Y|X)=P_U(Y|X)$, for the same reasons why $P(Y|X)$ is altered in prior probability shift. However, under mild conditions, we can assume $P(Y|X)$ does not change, or does not change significantly.}


\subsection{Concept Shift}
\label{sec:conceptshift}

\begin{figure}[t!]
 \begin{center}
 \includegraphics[width=\textwidth]{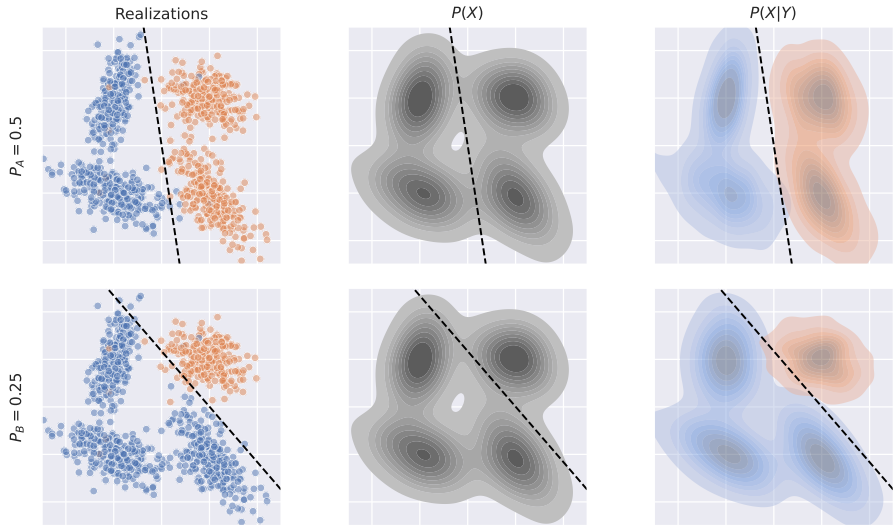}
 \end{center}
 \caption{Example of concept shift generated with synthetic data
 using a normal distribution for each cluster. Situation (a) (1st
 row): original data distribution. Situation (b) (2nd row): the
 concept ``negative'' (blue) has changed in a way that it now
 encompasses one of the originally ``positive'' (orange) clusters,
 thus implying a change in $P(X|Y)$ and in $P(Y|X)$ but not in $P(X)$
 (2nd column).}
 \label{fig:concept}
\end{figure}

\noindent Concept shift arises when the boundaries of the classes
change, i.e., when the underlying \emph{concepts} of interest change
across the training and the testing conditions. Concept shift is
characterised by a change in the class-conditional distribution
$P_{L}(X|Y) \neq P_{U}(X|Y)$, as well as a change in the posterior
distribution $P_{L}(Y|X) \neq P_{U}(Y|X)$. Another way of saying this
is that there is a change in the functional relationship between the
covariates and the class labels; see Figure~\ref{fig:concept}.

Figure~\ref{fig:concept} depicts a situation in which each of the two
classes (say, documents relevant and non relevant, respectively, to a
certain user information need) subsumes two subclasses, and one of the
subclasses ``switches class'', i.e., the documents contained in the
subclass were once considered relevant to the information need and are
now not relevant any more. Yet another example along these lines could
be due to a change in the sensitivity of a response variable. So, for
example, a change in the threshold above which the value of a
continuous response variable indicates a positive example, is a change
in the concept of ``being positive'', which implies (i) a change in
$P(Y|X)$, since some among the positive examples have now become
negative, (ii) a change in $P(X|Y)$, since the positive and negative
classes are inevitably distributed differently, and (iii) even a
change in $P(Y)$, since the higher the threshold, the fewer the
positive examples; however, the above does not imply any change in the
marginal distribution $P(X)$.

There are other examples of concept shift which may, instead, lead to
a change in $P(X)$ as well. Take, for example, the case of
epidemiology (one of the quintessential applications of
quantification) in which the spread of a disease (e.g., by a viral
infection) is now manifested in the population by means of different
symptoms (the covariates) due to a change in the pathogenic source
(e.g., a mutation). In this paper, though, we will only be
considering instances of concept shift in which the marginal
distribution $P(X)$ does not change, since otherwise none of the four
distributions of interest ($P(X)$, $P(Y)$, $P(X|Y)$, $P(Y|X)$) would
be invariant across $L$ and $U$, which would make the problem
essentially unlearnable.



Needless to say, concept shift represents the hardest type of shift
for any quantification system (and, more in general, for any inductive
inference model), since changes in the concept being modelled are
external to the learning procedure, and since there is no possibility
of behaving robustly to arbitrary changes in the functional
relationship between the covariates and the labels. Attempts to tackle
concept shift should inevitably entail a later phase of learning (as
in continuous learning) in which the model is informed, possibly by
means of new labelled examples,
of the changes in the functional relationship between covariates and
classes. To date, we are unaware of the existence of quantification
methods devised to counter concept shift.

\subsection{Recapitulation}
\label{sec:recap}

\noindent In light of the considerations above, in
Table~\ref{tab:oursetting} we present the specific types of shift that
we consider in this paper. Concretely, this comes down to exploring
plausible ways of filling out the blank cells of
Table~\ref{tab:datasetshift}, which are indicated in grey in
Table~\ref{tab:oursetting}.

\begin{table}[t]
 \resizebox{\textwidth}{!} {
 \begin{tabular}{l|c||c|c|c||l|c}
 & $P(X)$ & $P(Y)$ & $P(X|Y)$ & $P(Y|X)$ & Definition & Experiments \\ 
 \hline
 Prior probability shift & \gray $\neq$ & $\neq$ & = & \gray 
 $\neq$ & 
 \S\ref{sec:priorprobabilityshift} & \S\ref{sec:exp:prior} \\
 \hline
 Global pure covariate shift & $\neq$ & \gray = & \gray $\neq$ & = & 
 \S\ref{sec:globalcovariateshift} & \S\ref{sec:exp:covariate} \\ 
 \hline
 Global mixed covariate shift & $\neq$ & \gray $\neq$ & \gray 
 $\neq$ & = & 
 \S\ref{sec:globalcovariateshift} & \S\ref{sec:exp:covariate} \\
 \hline
 Local covariate shift & $\neq$ & \gray $\neq$ & \gray 
 $\neq$ & \phantom{$^*$}=$^{\blue{*}}$ &
 \S\ref{sec:localcovariateshift} & \S\ref{sec:exp:changes}\\
 \hline
 Concept shift & \gray = & \gray $\neq$ & $\neq$ & 
 $\neq$ 
 & \S\ref{sec:conceptshift} & \S\ref{sec:exp:concept}\\
 \hline
 \end{tabular}
 }
\caption{The types of shift we consider. Greyed-out cells indicate assumptions we make (and that we discuss and justify in Section~\ref{sec:shift}). \blue{Symbol * indicates a condition that can get compromised in extreme situations.}}
\label{tab:oursetting}
\end{table}


\section{Experiments}
\label{sec:experiments}

\noindent In this section we describe experiments that we have carried
out in which we simulate the different types of dataset shift
described in the previous sections. For simplicity, we have simulated
all these types of shift by using the same base datasets, which we
describe in the following section.


\subsection{Datasets}
\label{sec:dataset}

\noindent We extract the datasets we use for the experiments from a
large crawl of 233.1M Amazon product reviews made available by
\citet{McAuley:2015ss};\footnote{\url{http://jmcauley.ucsd.edu/data/amazon/links.html}}
we use different datasets for simulating different types of shift.
\alexcomment{Zenodo link?}
In order to extract these datasets from this crawl we first remove (a)
all product reviews shorter than 200 characters and (b) all product
reviews that have not been recognised as ``useful'' by any users. We
concentrate our attention on two merchandise categories,
\textsc{Books} and \textsc{Electronics}, since these are the two most
populated categories in the corpus; in the next sections these two
categories will sometimes be referred to as category $A$ and category
$B$.

Every review comes with a (true) label, consisting of the number of
stars (according to a ``5-star rating'', with 1 star standing for
``poor'' and 5 stars standing for ``excellent'') that the author
herself has attributed to the product being reviewed. Note that the
classes are ordered, and thus we can define
$\mathcal{Y}_{\star}=\{s_{1}, s_{2}, s_{3}, s_{4}, s_{5}\}$, with
$s_{i}$ meaning ``$i$ stars'', and
$s_{1} \prec s_{2} \prec s_{3} \prec s_{4} \prec s_{5}$. Since we deal
with binary quantification, we exploit this order to generate, at
desired ``cut points'' (i.e., thresholds below which a review is
considered negative and above which is considered positive), binary
versions of the dataset.
We thus define the function ``$\operatorname{binarise\_dataset}$'',
that takes a dataset labelled according to $\mathcal{Y}_{\star}$ and a
cut point $c$, and returns a new version of the dataset labelled
according to a binary codeframe $\mathcal{Y}=\{0, 1\}$; here, every
labelled datapoint $(\mathbf{x}, s_i)$, with
$s_i\in\mathcal{Y}_{\star}$, is converted into a datapoint
$(\mathbf{x}, y)$, with $y\in\mathcal{Y}$, such that $y=1$ (the
positive class) if $i>c$, or $y=0$ (the negative class) if $i<c$; note
that we filter out datapoints for which $i=c$. In the cases in which
we want to retain all datapoints labelled with all possible numbers of
stars, we simply specify $c$ as a real value intermediate between two
integers (e.g., $c=2.5$).


%
\begin{table}[t!]
 \centering \scalebox{1}{ \begin{tabular}{lllllll}
\toprule
 & instances & $\filledstar\smallstar\smallstar\smallstar\smallstar$ & $\filledstar\filledstar\smallstar\smallstar\smallstar$ & $\filledstar\filledstar\filledstar\smallstar\smallstar$ & $\filledstar\filledstar\filledstar\filledstar\smallstar$ & $\filledstar\filledstar\filledstar\filledstar\filledstar$ \\
\midrule
\textsc{Books} & 7,813,813 & 0.093 & 0.071 & 0.094 & 0.160 & 0.582 \\
\textsc{Electronics} & 1,889,965 & 0.193 & 0.079 & 0.093 & 0.178 & 0.457 \\
\bottomrule
\end{tabular}
}
 \label{tab:datasetinfo}
 \caption{Dataset information for categories \textsc{Books} and
 \textsc{Electronics}, along with the prevalence for each different
 star rating.}
\end{table}


\subsection{General Experimental Setup}
\label{sec:generalexpsetup}

\noindent In all the experiments carried out in this study we fix the
size of the training set to 5,000 and the size of each test sample to
500. For a given experiment we evaluate all quantification methods
with the same test samples, but different experiments may involve
different samples depending on the type of shift being simulated. We
run different experiments, each targeting a specific type of dataset
shift; within each experiment we simulate the presence, in a
systematic and controlled manner, of different degrees of shift. When
testing with different degrees of a given type of shift, for every
such degree we randomly generate 50 test samples. In order to account
for stochastic fluctuations in the results due to the random selection
of a particular training set, we repeat each experiment 10 times.
%
%
%
%
%
%
We carry out all the experiments by using the QuaPy open-source
quantification library
\citep{Moreo:2021bs}.\footnote{\url{https://github.com/HLT-ISTI/QuaPy}}
All the code for reproducing our experiments is available from a
dedicated GitHub
repository.\footnote{\url{https://github.com/pglez82/quant_datasetshift}}

In order to turn raw documents into vectors, as the features we use
tfidf-weighted words; we compute idf independently for each experiment
by only taking into account the 5,000 training documents selected for
that experiment. We only retain the words appearing at least 3 times
in the training set, meaning that the number of different words
(hence, the number of dimensions in the vector space) can vary across
experiments.

As the evaluation measure we use absolute error (AE), since it is one
of the most satisfactory (see~\citep{Sebastiani:2020qf} for a
discussion) and frequently used measures in quantification
experiments, and since it is very easily interpretable. In the binary
case, AE is defined as
%
%
\begin{align}
 \operatorname{AE}(p_{\sigma},\hat{p}_{\sigma})= |p_{\sigma}-\hat{p}_{\sigma}|
\end{align}
%
%
For each experiment we report the mean absolute error (MAE), where the
mean is computed across all the samples with the same degree of shift
and all the repetitions thereof. We perform statistical significance
tests at different confidence levels in order to check for the
differences in performance between the best method (highlighted in
boldface in all tables) and all other competing methods. All methods
whose scores are \emph{not} statistically significantly different from
the best one, according to a Wilcoxon signed-rank test on paired
samples, are marked with a special symbol. We use superscript $\dag$
to indicate that $0.001 <$ \textit{p}-value $< 0.05$ (loosely
speaking, that the scores are ``somewhat similar''),
while superscript $\ddag$ indicates that $0.05\leq$ \textit{p}-value
(loosely speaking, that the scores are ``very similar'');
the absence of any such symbol thus indicates that \textit{p}-value
$\leq 0.001$ (loosely speaking, that the scores are ``fairly
different'').

All the quantification methods considered in this study are of the
aggregative type and are described in Section~\ref{sec:methods}. In
addition to these methods, we had initially also considered the Sample
Mean Matching (SMM) method~\citep{Hassan:2020kq}, but we removed this
method from the experiments as we found it to be equivalent to the
PACC method (we give a formal proof of this equivalence in
Appendix~\ref{sec:smmpacceq}).

For the sake of fairness, underlying all quantification methods we use
the same type of classifier. (All the quantification methods we use
are aggregative, so all of them use an underlying classifier.)
As our classifier of choice we use logistic regression, since it is a
well-known classifier which also delivers ``soft'' predictions and is
known to deliver reasonably well-calibrated posterior probabilities
(these two characteristics are required for PCC, PACC, DyS, and
SLD). We optimise the hyperparameters of the quantifier following
\cite{Moreo:2021sp}, i.e., minimising a quantification-oriented loss
function (here: MAE) via a quantification-oriented parameter
optimisation protocol; we explore the values
$C\in\{0.1,1,10,100,100\}$ (where $C$ is the inverse of the
regularization strength), and the values \textsc{class\_weight} $\in$
$\{\operatorname{Balanced}$, $\operatorname{None}\}$ (where
\textsc{class\_weight} indicates the relative importance of each
class), via grid search. We evaluate each configuration of
hyperparameters in terms of MAE over artificially generated samples
using a held-out stratified validation set consisting of 40\% of the
training documents. This means that we optimise each classifier
specifically for each quantifier, and the parameters we choose are the
ones that best suit this particular quantifier. Once we have chosen
the optimal values for the hyperparameters, we retrain the quantifier
using the entire training set.

The quantification methods used in this study do not have any
additional hyperparameters, except for DyS that has two, i.e., (i) the
number of bins used to build the histograms and (ii) the distance
function. In this work we fix these values to (i) 10 bins and (ii) the
Topsoe distance, since these are the values that gave the best results
in the work that originally introduced DyS~\citep{Maletzke:2019qd}.


\subsection{Prior Probability Shift}
\label{sec:exp:prior}


\subsubsection{Evaluation Protocol}
\label{sec:exp:prior:prot}

\noindent For generating prior probability shift we consider all the
reviews from categories \textsc{Electronics} and
\textsc{Books}. Algorithm~\ref{alg:prior} describes the experimental
setup for this type of shift. For binarising the dataset we follow the
approach described in Section~\ref{sec:dataset}, using a cut point of
$3$. We sample 5,000 training documents from the dataset using
prevalence values of the positive class with values ranging from $0$
to $1$, at steps of $0.1$. (Since it is not possible to generate a
classifier with no positive examples or no negative examples, we
actually replace $p_{L}=0$ and $p_{L}=1$ with $p_{L}=0.02$ and
$p_{L}=0.98$, respectively.) We draw test samples from the dataset
varying, here too, the prevalence of the positive class using values
in $\{0.0,0.1,..,0.9,1.0\}$. In order to give a quantitative
indication of the degree of prior probability shift in each
experiment, we compute the
signed difference $(p_{L}-p_{U})$ rounded to one decimal, resulting in
a real value in the range $[-1,1]$; to this respect, note that
negative degrees of shift do \textit{not} indicate an absence of
shift, but indicate a presence of shift in which $p_{U}$ is greater
than $p_{L}$ (for positive degrees, $p_{U}$ is lower than $p_{L}$).

For this experiment the number of test samples used for evaluation
amounts to $11\times11\times50\times10$ = 60,500 for each
quantification algorithm we test.

\begin{algorithm}[t!]
 \hspace*{\algorithmicindent} \textbf{Input:} Datasets $A$ and $B$;
 Quantification learner $Q$ \\
 \begin{algorithmic}[1]

 \State $D \gets A \cup B$ \State
 $D \gets \operatorname{binarise\_dataset}(D,
 \operatorname{cut\_point}=3)$ \State
 $\mathcal{L},\; \mathcal{U} \gets
 \operatorname{split\_stratified(D)}$ \For{10 repetitions}
 \For{$p^{L} \in \{0.02,0.1,0.2,...,0.8,0.9,0.98\}$} \State
 {\commentfont /* Generate a sample from $\mathcal{L}$ with
 prevalence $p^{L}$ */} \State $L \sim \mathcal{L}$ with
 $p_{L}=p^{L}$ and $|L|=5000$

 \State{\commentfont /* Use algorithm $Q$ to learn a quantifier $q$
 on $L$ */} \State $q \gets Q.fit(L)$ \For{50 repetitions}
 \State{\commentfont /* Generating test samples */}
 \For{$p^{U} \in \{0.0,0.1, ...,0.9,1.0\}$}

 \State $U \sim \mathcal{U}$ with $p_{U}=p^{U}$ and $|U|=500$

 \State $\hat{p}_{U}^{q} \gets q.\operatorname{quantify}(U)$

 \State $error \gets \operatorname{AE}(p_{U}, \hat{p}_{U}^{q})$
 \EndFor \EndFor \EndFor \EndFor
 \end{algorithmic}
 \caption{Protocol for generating prior probability shift.}
 \label{alg:prior}
\end{algorithm}


\subsubsection{Results}
\label{sec:resultsprior}

\begin{figure}[t!]
 \centering
 \includegraphics[width=\textwidth]{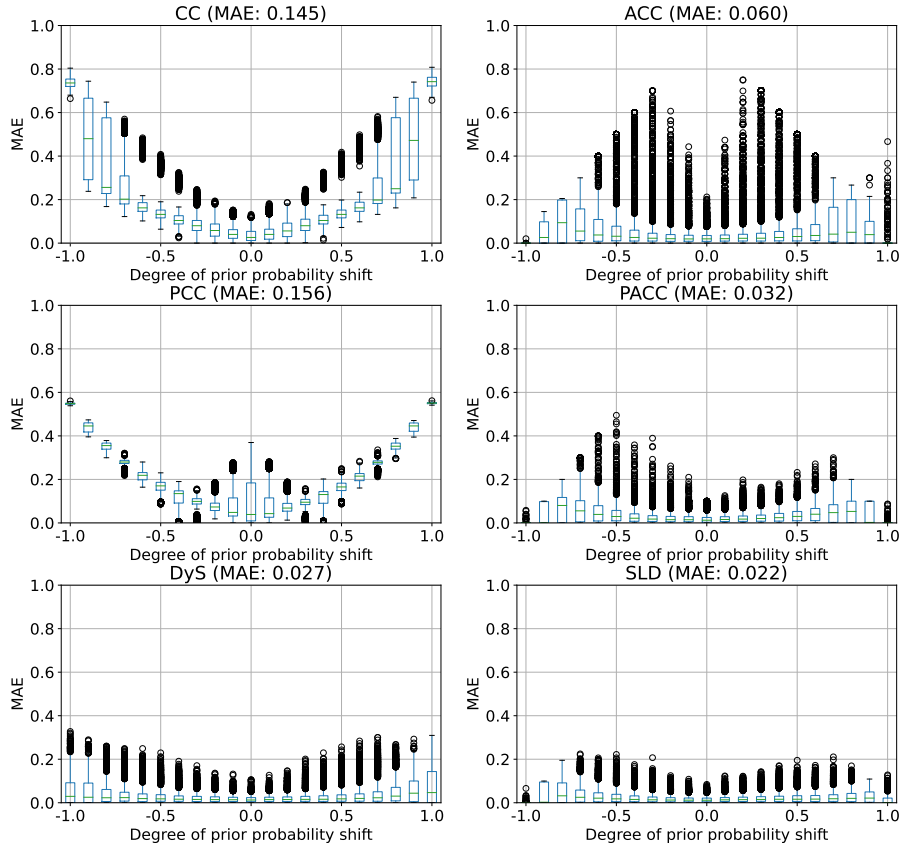}
 \caption{Results obtained for prior probability shift; the error
 measure is MAE and the degree of shift is computed as
 $(p_{U}-p_{L})$ (rounded to one decimal). }
 \label{fig:priorresults}
\end{figure}

\noindent Table~\ref{tab:priortab} and Figure~\ref{fig:priorresults}
present the results of the prior probability shift experiments in the
form of boxplots (blue boxes), where the outliers are indicated by
black dots. In this case the SLD method stands out as the best
performer, closely followed by DyS and PACC. These methods perform
very well when the degree of shift is moderate,\footnote{Here and in
the rest of the paper, when speaking of ``high'' or ``low'' degrees of
shift we actually refer to the absolute value of this degree (e.g., a
degree of shift of -1 counts as as a ``high'' degree of shift). This
will be the case not only for prior probability shift but also for
other types of shift.} while their performance degrades as this degree
increases. On the other hand, CC and PCC are clearly the worst
performers; the reason is that, as stated previously, CC and PCC
naturally inherit the bias of the underlying classifier, so when the
divergence between the distribution they are biased towards (i.e., the
training distribution) and the test distribution increases, their
performance tends to decrease. These results are in line with previous
studies in the quantification literature (e.g.,
\citep{Maletzke:2019qd, Schumacher:2021ty, Moreo:2021bs,
Moreo:2022bf}), most of which has indeed focused on prior probability
shift.


\begin{table}[t!]
 \centering \begin{tabular}{r|rrrrrr}
\toprule
 & CC & ACC & PCC & PACC & DyS & SLD \\
\midrule
-1.0 & .737 & \textbf{.000} & .548 & .001 & .063 & .001 \\
-0.9 & .479 & .049 & .439 & .044 & \ddag{.053} & \textbf{.041} \\
-0.8 & .355 & .088 & .352 & .077 & \textbf{.045} & .049 \\
-0.7 & .271 & .099 & .278 & .069 & \textbf{.040} & \ddag{.041} \\
-0.6 & .213 & .094 & .216 & .054 & \textbf{.032} & \dag{.034} \\
-0.5 & .166 & .086 & .162 & .042 & \textbf{.028} & \ddag{.029} \\
-0.4 & .126 & .071 & .115 & .031 & \dag{.024} & \textbf{.023} \\
-0.3 & .091 & .055 & .093 & .025 & .021 & \textbf{.020} \\
-0.2 & .064 & .041 & .085 & .023 & .019 & \textbf{.017} \\
-0.1 & .047 & .032 & .091 & .022 & .017 & \textbf{.015} \\
0.0 & .035 & .026 & .111 & .017 & .016 & \textbf{.014} \\
0.1 & .048 & .034 & .090 & .021 & .018 & \textbf{.017} \\
0.2 & .064 & .046 & .084 & .023 & \dag{.019} & \textbf{.018} \\
0.3 & .092 & .063 & .089 & .025 & .022 & \textbf{.020} \\
0.4 & .127 & .077 & .112 & .029 & .026 & \textbf{.022} \\
0.5 & .167 & .089 & .160 & .036 & .030 & \textbf{.024} \\
0.6 & .213 & .096 & .213 & .045 & .035 & \textbf{.025} \\
0.7 & .272 & .095 & .276 & .053 & .043 & \textbf{.027} \\
0.8 & .355 & .081 & .351 & .058 & .053 & \textbf{.030} \\
0.9 & .478 & .052 & .440 & .039 & .062 & \textbf{.029} \\
1.0 & .742 & .020 & .551 & \textbf{.002} & .076 & .016 \\
\bottomrule
\end{tabular}

 \caption{Results for prior probability shift experiments in terms of
 $\operatorname{MAE}$. Each row corresponds to a given degree of
 shift, measured as $(p_{U}-p_{L})$ (rounded to one decimal).}
 \label{tab:priortab}
\end{table}


One interesting observation that emerges from
Figure~\ref{fig:priorresults} has to do with the stability of the
methods. ACC shows a tendency to sporadically yield anomalously high
levels of error. Those levels of error correspond to cases in which
the training sample is severely imbalanced ($p_{L}=0.02$ or
$p_{L}=0.98$).
%
Note that, the correction implemented by
Equation~\ref{eq:acc2} may turn unreliable when the estimation of $\operatorname{tpr}$  itself is unreliable
(this is likely to occur when the amount of positives is 2\%, i.e., when $p_{L}=0.02$) and/or 
when the estimation of $\operatorname{fpr}$ is unreliable 
(this is likely to occur when the amount of negatives is 2\%, i.e., when $p_{L}=0.98$).
Yet another cause might include the instability of the denominator
(this happens when $\operatorname{tpr}\approx\operatorname{fpr}$), 
which could, in turn, require clipping the output in the range $[0,1]$. 
After analyzing the 100 worst cases, we verified that in 
36\% of the cases involved clipping,
in 46\% of the cases the denominator turned out to be smaller than 0.05.


Note that, if these extreme cases were to be removed, the average
scores obtained by ACC would not substantially differ from those
obtained by other quantification methods such as PACC or DyS.


\subsection{Global Covariate Shift}
\label{sec:exp:covariate}


\subsubsection{Evaluation Protocol}
\label{sec:exp:covariate:prot}

\noindent For generating global covariate shift, we modify the ratio
between the documents in category $A$ (\textsc{Books}) and those in
category $B$ (\textsc{Electronics}), across the training data and the
test samples. We binarise the dataset at a cut point of 3, as
described in Section~\ref{sec:dataset}. We vary the prevalence
$\alpha$ of category $A$ (the prevalence of category $B$ is
$(1-\alpha)$), in the training data ($\alpha^{L}$) and in the test
samples ($\alpha^{U}$), in the range $[0,1]$ with steps of 0.1, thus
giving rise to 121 possible combinations. For the sake of a clear
exposition, we present the results for different degrees of global
covariate shift, measured as the signed difference between
$\alpha^{L}$ and $\alpha^{U}$, resulting in a real value in the range
$[-1,+1]$. We vary the priors of the positive class using the values
$\{0.25, 0.50, 0.75\}$ in both the training data and the test samples,
in order to simulate cases of global pure covariate shift, where
$P_{L}(Y)=P_{U}(Y)$, and global mixed covariate shift, where
$P_{L}(Y) \neq P_{U}(Y)$. Note that even if the global pure covariate
shift scenario is particularly awkward for a quantification setting
(since the prevalence of the positive class in the training data
coincides with the one in the test data), it is interesting because it
shows how quantifiers react just to a mere change in the
covariates. Algorithm~\ref{alg:covariate} describes the experimental
setup for this type of shift.

For this experiment the number of test samples used for evaluation
amounts to $3\times3\times11\times11\times50\times10$ = 544,500 for
each quantification algorithm we test.

\begin{algorithm}[t!]
 \hspace*{\algorithmicindent} \textbf{Input:} Datasets $A$ and $B$;
 Quantification algorithm $Q$ \\
 \begin{algorithmic}[1]
 \State
 $A \gets \operatorname{binarise\_dataset}(A,
 \operatorname{cut\_point=3})$ \State
 $B \gets \operatorname{binarise\_dataset}(B,
 \operatorname{cut\_point=3})$ \State
 $\mathcal{L}_A, \mathcal{U}_A \gets
 \operatorname{split\_stratified}(A)$ \State
 $\mathcal{L}_B, \mathcal{U}_B \gets
 \operatorname{split\_stratified}(B)$ \For{10 repetitions}
 \For{$p^{L} \in \{0.25, 0.50, 0.75\}$}
 \For{$\alpha^{L} \in \{0.0,0.1,...,0.9,1.0\}$} \State
 {\commentfont /* Generate a training sample $L$ from
 $\mathcal{L}_A$ and $\mathcal{L}_B$ with prevalence
 $p^{L}$\\\hspace{1.4cm}and a proportion $\alpha^{L}$ of documents
 from $\mathcal{L}_A$ */}

 \State $L_A \sim \mathcal{L}_A$ with $p_{L_A}=p^{L}$ and
 $|L_A|=\lceil\alpha^{L} \cdot 5000\rceil$ \State
 $L_B \sim \mathcal{L}_B$ with $p_{L_B}=p^{L}$ and
 $|L_B|=\lfloor(1-\alpha^{L}) \cdot 5000\rfloor$ \State
 $L \gets L_A \cup L_B $

 \State {\commentfont /* Use quantification algorithm $Q$ to learn
 a quantifier $q$ on $L$ */} \State $q \gets Q.fit(L)$ \For{50
 repetitions} \State {\commentfont /* Generating test samples */}
 \For{$p^{U} \in \{0.25, 0.50, 0.75\}$}
 \For{$\alpha^{U} \in \{0.0,0.1,...,0.9,1.0\}$}

 \State $U_A \sim \mathcal{U}_A$ with $p_{U_A}=p^{U}$ and
 $|U_A|=\lceil\alpha^{U} \cdot 500\rceil$ \State
 $U_B \sim \mathcal{U}_B$ with $p_{U_B}=p^{U}$ and
 $|U_B|=\lfloor(1-\alpha^{U}) \cdot 500\rfloor$ \State
 $U \gets U_A \cup U_B $

 \State $\hat{p}_{U}^{q} \gets q.\operatorname{quantify}(U)$ \State
 $error \gets AE(p_{U},\hat{p}_{U}^{q})$ \EndFor \EndFor \EndFor
 \EndFor

 \EndFor \EndFor
 \end{algorithmic}
 \caption{Protocol for generating global covariate shift.}
 \label{alg:covariate}
\end{algorithm}


\subsubsection{Results}
\label{sec:exp:covariate:res}

\begin{figure}[t!]
 \includegraphics[width=\textwidth]{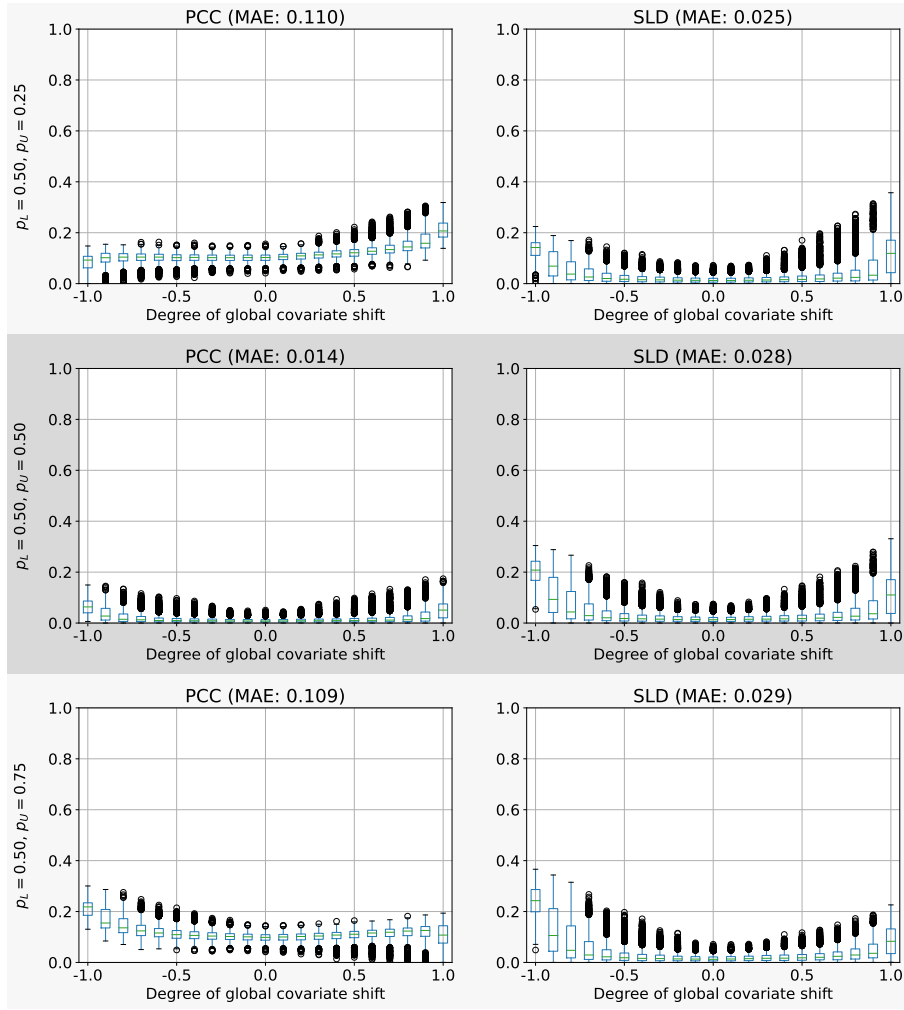}
 \caption{Results for global covariate shift with $p_{L}=0.5$. The
 error measure is MAE and the degree of covariate shift is computed
 as $(\alpha^{L}-\alpha^{U})$. Figures with a grey background
 represent cases of global \emph{pure} covariate shift, in which
 $P_{L}(Y)=P_{U}(Y)$.}
 \label{fig:covariate05}
\end{figure}

\begin{figure}[t!]
 \includegraphics[width=\textwidth]{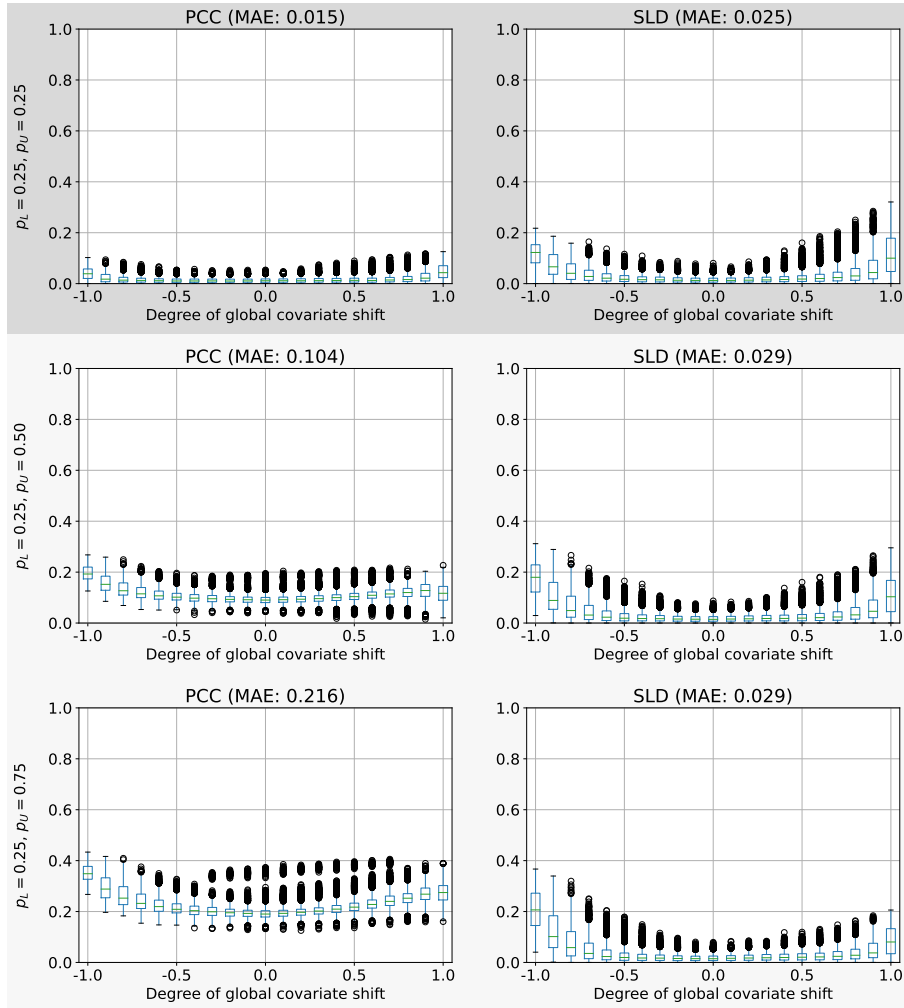}
 \caption{Results for global covariate shift with $p_{L}=0.25$. Error
 measure is MAE and the degree of covariate shift is computed as
 $(\alpha^{L}-\alpha^{U})$. Figures with a grey background represent
 cases of global \emph{pure} covariate shift, in which
 $P_{L}(Y)=P_{U}(Y)$.}
 \label{fig:covariate025}
\end{figure}

\begin{figure}[t!]
 \includegraphics[width=\textwidth]{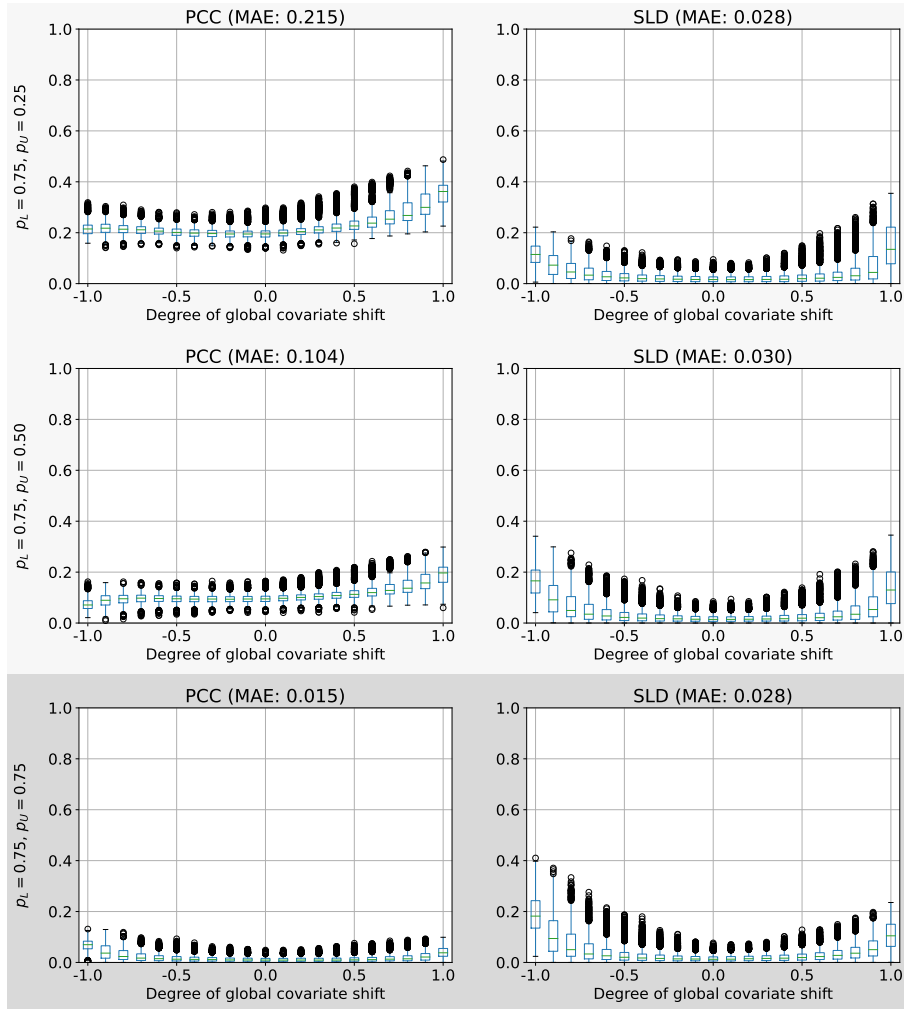}
 \caption{Results for global covariate shift with $p_{L}=0.75$. Error
 measure is MAE and the degree of covariate shift is computed as
 $(\alpha^{L}-\alpha^{U})$. Figures with a grey background represent
 cases of global \emph{pure} covariate shift, in which
 $P_{L}(Y)=P_{U}(Y)$.}
 \label{fig:covariate075}
\end{figure}

\noindent We now report the results for the scenario in which the data
exhibits global \textit{pure} covariate shift (see Tables
\ref{tab:covariate05}, \ref{tab:covariate025}, \ref{tab:covariate075},
where global pure covariate shift is represented by the columns with a
grey background, and Figures \ref{fig:covariate05},
\ref{fig:covariate025}, \ref{fig:covariate075}). As can be expected,
the bigger the degree of such shift, the worse the performance of the
methods. Note that a degree of global pure covariate shift equal to 1
(resp., -1) means that the system was trained with documents only from
category $A$ (resp., $B$) while the testing samples only have
documents from category $B$ (resp., $A$). On the other hand, low
degrees of global pure covariate shift represent the situation in
which similar values of $\alpha^L$ and $\alpha_U$ were used. The
experiments show that the method most robust to global pure covariate
shift
is PCC, which is consistent with the theoretical results of
\citet{Tasche:2022fu}. PCC is able to provide good results, beating
the other methods consistently, even when the degree of global pure
covariate shift is high. On the other hand, methods like SLD, that
show excellent performance under prior probability shift, perform
poorly under high values of global pure covariate shift.

\begin{table}[t!]
 \setlength{\tabcolsep}{2.5pt} {\resizebox{\textwidth}{!}{
 \begin{tabular}{r|rrrrrr|rrrrrr|rrrrrr}
\toprule
 & \multicolumn{6}{c}{$p_{U}=0.25$} & \multicolumn{6}{c}{$p_{U}=0.5$} & \multicolumn{6}{c}{$p_{U}=0.75$} \\
 & CC & ACC & PCC & PACC & DyS & SLD & CC & ACC & PCC & PACC & DyS & SLD & CC & ACC & PCC & PACC & DyS & SLD \\
\midrule
-1.0 & \textbf{.029} & .069 & .085 & .044 & .080 & .133 & \cellcolor{verylightgray} .107 & \cellcolor{verylightgray} .147 & \textbf{\cellcolor{verylightgray} .065} & \cellcolor{verylightgray} .109 & \cellcolor{verylightgray} .142 & \cellcolor{verylightgray} .200 & .231 & .225 & .213 & \ddag{.190} & \textbf{.188} & .236 \\
-0.9 & .052 & .046 & .098 & \textbf{.033} & .052 & .078 & \cellcolor{verylightgray} .061 & \cellcolor{verylightgray} .083 & \textbf{\cellcolor{verylightgray} .037} & \cellcolor{verylightgray} .064 & \cellcolor{verylightgray} .084 & \cellcolor{verylightgray} .110 & .166 & .132 & .169 & \ddag{.114} & \textbf{.112} & .128 \\
-0.8 & .060 & .034 & .101 & \textbf{.025} & .037 & .052 & \cellcolor{verylightgray} .042 & \cellcolor{verylightgray} .057 & \textbf{\cellcolor{verylightgray} .025} & \cellcolor{verylightgray} .043 & \cellcolor{verylightgray} .055 & \cellcolor{verylightgray} .071 & .137 & .090 & .145 & \dag{.076} & \textbf{.074} & .082 \\
-0.7 & .065 & .029 & .102 & \textbf{.023} & .031 & .039 & \cellcolor{verylightgray} .033 & \cellcolor{verylightgray} .044 & \textbf{\cellcolor{verylightgray} .020} & \cellcolor{verylightgray} .035 & \cellcolor{verylightgray} .043 & \cellcolor{verylightgray} .051 & .117 & .064 & .130 & \ddag{.054} & \textbf{.053} & \ddag{.055} \\
-0.6 & .068 & .025 & .103 & \textbf{.021} & .025 & .030 & \cellcolor{verylightgray} .025 & \cellcolor{verylightgray} .034 & \textbf{\cellcolor{verylightgray} .015} & \cellcolor{verylightgray} .027 & \cellcolor{verylightgray} .033 & \cellcolor{verylightgray} .037 & .103 & .047 & .120 & \textbf{.039} & \ddag{.039} & \ddag{.040} \\
-0.5 & .070 & .022 & .101 & \textbf{.020} & .022 & .024 & \cellcolor{verylightgray} .021 & \cellcolor{verylightgray} .029 & \textbf{\cellcolor{verylightgray} .014} & \cellcolor{verylightgray} .024 & \cellcolor{verylightgray} .028 & \cellcolor{verylightgray} .030 & .093 & .038 & .113 & .032 & .032 & \textbf{.031} \\
-0.4 & .072 & \dag{.021} & .102 & \ddag{.020} & \textbf{.020} & \ddag{.020} & \cellcolor{verylightgray} .018 & \cellcolor{verylightgray} .025 & \textbf{\cellcolor{verylightgray} .012} & \cellcolor{verylightgray} .021 & \cellcolor{verylightgray} .024 & \cellcolor{verylightgray} .025 & .085 & .032 & .109 & .029 & .026 & \textbf{.025} \\
-0.3 & .074 & .020 & .101 & .020 & \ddag{.019} & \textbf{.018} & \cellcolor{verylightgray} .016 & \cellcolor{verylightgray} .023 & \textbf{\cellcolor{verylightgray} .011} & \cellcolor{verylightgray} .019 & \cellcolor{verylightgray} .021 & \cellcolor{verylightgray} .021 & .080 & .027 & .105 & .025 & .022 & \textbf{.021} \\
-0.2 & .074 & .019 & .101 & .019 & .018 & \textbf{.016} & \cellcolor{verylightgray} .015 & \cellcolor{verylightgray} .020 & \textbf{\cellcolor{verylightgray} .010} & \cellcolor{verylightgray} .017 & \cellcolor{verylightgray} .019 & \cellcolor{verylightgray} .018 & .075 & .023 & .102 & .022 & .020 & \textbf{.018} \\
-0.1 & .076 & .019 & .101 & .019 & .017 & \textbf{.015} & \cellcolor{verylightgray} .014 & \cellcolor{verylightgray} .019 & \textbf{\cellcolor{verylightgray} .010} & \cellcolor{verylightgray} .017 & \cellcolor{verylightgray} .018 & \cellcolor{verylightgray} .016 & .072 & .021 & .100 & .020 & .018 & \textbf{.016} \\
0.0 & .077 & .018 & .102 & .017 & .017 & \textbf{.015} & \cellcolor{verylightgray} .014 & \cellcolor{verylightgray} .018 & \textbf{\cellcolor{verylightgray} .009} & \cellcolor{verylightgray} .016 & \cellcolor{verylightgray} .017 & \cellcolor{verylightgray} .015 & .068 & .019 & .098 & .018 & .017 & \textbf{.015} \\
0.1 & .080 & .019 & .104 & .017 & .018 & \textbf{.015} & \cellcolor{verylightgray} .014 & \cellcolor{verylightgray} .019 & \textbf{\cellcolor{verylightgray} .010} & \cellcolor{verylightgray} .016 & \cellcolor{verylightgray} .018 & \cellcolor{verylightgray} .016 & .068 & .019 & .099 & .019 & .018 & \textbf{.016} \\
0.2 & .084 & .022 & .108 & .020 & .020 & \textbf{.016} & \cellcolor{verylightgray} .015 & \cellcolor{verylightgray} .020 & \textbf{\cellcolor{verylightgray} .011} & \cellcolor{verylightgray} .018 & \cellcolor{verylightgray} .020 & \cellcolor{verylightgray} .017 & .068 & .020 & .100 & .019 & .019 & \textbf{.017} \\
0.3 & .087 & .025 & .112 & .023 & .024 & \textbf{.017} & \cellcolor{verylightgray} .016 & \cellcolor{verylightgray} .021 & \textbf{\cellcolor{verylightgray} .011} & \cellcolor{verylightgray} .019 & \cellcolor{verylightgray} .023 & \cellcolor{verylightgray} .019 & .069 & .020 & .102 & \textbf{.019} & .020 & \ddag{.019} \\
0.4 & .092 & .030 & .117 & .028 & .028 & \textbf{.020} & \cellcolor{verylightgray} .018 & \cellcolor{verylightgray} .023 & \textbf{\cellcolor{verylightgray} .012} & \cellcolor{verylightgray} .020 & \cellcolor{verylightgray} .026 & \cellcolor{verylightgray} .021 & .070 & .021 & .105 & \textbf{.020} & .022 & \dag{.021} \\
0.5 & .097 & .035 & .122 & .035 & .033 & \textbf{.023} & \cellcolor{verylightgray} .019 & \cellcolor{verylightgray} .025 & \textbf{\cellcolor{verylightgray} .013} & \cellcolor{verylightgray} .022 & \cellcolor{verylightgray} .030 & \cellcolor{verylightgray} .023 & .073 & \ddag{.022} & .108 & \textbf{.022} & .025 & \ddag{.022} \\
0.6 & .105 & .044 & .130 & .045 & .042 & \textbf{.027} & \cellcolor{verylightgray} .022 & \cellcolor{verylightgray} .028 & \textbf{\cellcolor{verylightgray} .015} & \cellcolor{verylightgray} .025 & \cellcolor{verylightgray} .037 & \cellcolor{verylightgray} .027 & .075 & \textbf{.025} & .112 & .026 & .030 & \ddag{.026} \\
0.7 & .114 & .056 & .139 & .060 & .053 & \textbf{.034} & \cellcolor{verylightgray} .025 & \cellcolor{verylightgray} .032 & \textbf{\cellcolor{verylightgray} .018} & \cellcolor{verylightgray} .029 & \cellcolor{verylightgray} .046 & \cellcolor{verylightgray} .033 & .077 & \textbf{.028} & .114 & .032 & .036 & .030 \\
0.8 & .127 & .075 & .152 & .082 & .074 & \textbf{.044} & \cellcolor{verylightgray} .031 & \cellcolor{verylightgray} .040 & \textbf{\cellcolor{verylightgray} .022} & \cellcolor{verylightgray} .036 & \cellcolor{verylightgray} .061 & \cellcolor{verylightgray} .044 & .078 & \textbf{.033} & .117 & .041 & .046 & .039 \\
0.9 & .148 & .105 & .170 & .115 & .112 & \textbf{.063} & \cellcolor{verylightgray} .040 & \cellcolor{verylightgray} .052 & \textbf{\cellcolor{verylightgray} .030} & \cellcolor{verylightgray} .047 & \cellcolor{verylightgray} .089 & \cellcolor{verylightgray} .061 & .078 & \textbf{.041} & .118 & .052 & .060 & .050 \\
1.0 & .194 & .168 & .212 & .183 & .203 & \textbf{.124} & \cellcolor{verylightgray} .070 & \cellcolor{verylightgray} .090 & \textbf{\cellcolor{verylightgray} .055} & \cellcolor{verylightgray} .087 & \cellcolor{verylightgray} .155 & \cellcolor{verylightgray} .115 & .066 & \textbf{.053} & .107 & \dag{.056} & .093 & .086 \\
\bottomrule
\end{tabular}
 }}
 \caption{\label{tab:covariate05}Results for global covariate shift
 when $p_{L} = 0.5$ in terms of MAE. Each row contains the results
 for a degree in covariate shift computed as
 $(\alpha^{L}-\alpha^{U})$. Results are presented in three groups,
 depending on the prevalence used for the positive class in the test
 samples $p_{U}$. Columns with a grey background represent cases of
 of global \emph{pure} covariate shift, in which
 $P_{L}(Y)=P_{U}(Y)$.}
\end{table}

\begin{table}[t!]
 \setlength{\tabcolsep}{2.5pt}{\resizebox{\textwidth}{!}{
 \begin{tabular}{r|rrrrrr|rrrrrr|rrrrrr}
\toprule
 & \multicolumn{6}{c}{$p_{U}=0.25$} & \multicolumn{6}{c}{$p_{U}=0.5$} & \multicolumn{6}{c}{$p_{U}=0.75$} \\
 & CC & ACC & PCC & PACC & DyS & SLD & CC & ACC & PCC & PACC & DyS & SLD & CC & ACC & PCC & PACC & DyS & SLD \\
\midrule
-1.0 & \cellcolor{verylightgray} .059 & \cellcolor{verylightgray} .083 & \textbf{\cellcolor{verylightgray} .040} & \cellcolor{verylightgray} .067 & \cellcolor{verylightgray} .127 & \cellcolor{verylightgray} .115 & .200 & .175 & .195 & \textbf{.160} & .221 & .174 & .340 & .267 & .351 & .253 & .290 & \textbf{.207} \\
-0.9 & \cellcolor{verylightgray} .035 & \cellcolor{verylightgray} .053 & \textbf{\cellcolor{verylightgray} .024} & \cellcolor{verylightgray} .043 & \cellcolor{verylightgray} .081 & \cellcolor{verylightgray} .075 & .145 & .107 & .158 & \textbf{.097} & .138 & .107 & .265 & .168 & .296 & .158 & .180 & \textbf{.123} \\
-0.8 & \cellcolor{verylightgray} .029 & \cellcolor{verylightgray} .038 & \textbf{\cellcolor{verylightgray} .018} & \cellcolor{verylightgray} .032 & \cellcolor{verylightgray} .060 & \cellcolor{verylightgray} .050 & .114 & .070 & .136 & \textbf{.063} & .096 & .068 & .221 & .111 & .265 & .103 & .126 & \textbf{.080} \\
-0.7 & \cellcolor{verylightgray} .025 & \cellcolor{verylightgray} .031 & \textbf{\cellcolor{verylightgray} .016} & \cellcolor{verylightgray} .027 & \cellcolor{verylightgray} .044 & \cellcolor{verylightgray} .037 & .096 & .053 & .122 & \textbf{.048} & .069 & \ddag{.049} & .195 & .078 & .242 & .073 & .090 & \textbf{.054} \\
-0.6 & \cellcolor{verylightgray} .024 & \cellcolor{verylightgray} .026 & \textbf{\cellcolor{verylightgray} .014} & \cellcolor{verylightgray} .022 & \cellcolor{verylightgray} .034 & \cellcolor{verylightgray} .029 & .083 & .041 & .112 & .038 & .051 & \textbf{.035} & .175 & .058 & .226 & .056 & .065 & \textbf{.038} \\
-0.5 & \cellcolor{verylightgray} .023 & \cellcolor{verylightgray} .024 & \textbf{\cellcolor{verylightgray} .013} & \cellcolor{verylightgray} .020 & \cellcolor{verylightgray} .028 & \cellcolor{verylightgray} .023 & .075 & .035 & .106 & .033 & .041 & \textbf{.028} & .163 & .048 & .215 & .048 & .051 & \textbf{.029} \\
-0.4 & \cellcolor{verylightgray} .022 & \cellcolor{verylightgray} .023 & \textbf{\cellcolor{verylightgray} .012} & \cellcolor{verylightgray} .019 & \cellcolor{verylightgray} .024 & \cellcolor{verylightgray} .019 & .070 & .032 & .100 & .030 & .034 & \textbf{.024} & .154 & .042 & .207 & .045 & .040 & \textbf{.024} \\
-0.3 & \cellcolor{verylightgray} .022 & \cellcolor{verylightgray} .022 & \textbf{\cellcolor{verylightgray} .012} & \cellcolor{verylightgray} .018 & \cellcolor{verylightgray} .022 & \cellcolor{verylightgray} .018 & .065 & .030 & .098 & .029 & .028 & \textbf{.021} & .146 & .039 & .203 & .043 & .033 & \textbf{.021} \\
-0.2 & \cellcolor{verylightgray} .022 & \cellcolor{verylightgray} .021 & \textbf{\cellcolor{verylightgray} .012} & \cellcolor{verylightgray} .017 & \cellcolor{verylightgray} .020 & \cellcolor{verylightgray} .016 & .062 & .027 & .094 & .026 & .024 & \textbf{.019} & .142 & .034 & .197 & .039 & .028 & \textbf{.018} \\
-0.1 & \cellcolor{verylightgray} .021 & \cellcolor{verylightgray} .021 & \textbf{\cellcolor{verylightgray} .012} & \cellcolor{verylightgray} .017 & \cellcolor{verylightgray} .018 & \cellcolor{verylightgray} .015 & .061 & .026 & .093 & .025 & .021 & \textbf{.017} & .140 & .031 & .195 & .036 & .023 & \textbf{.017} \\
0.0 & \cellcolor{verylightgray} .021 & \cellcolor{verylightgray} .021 & \textbf{\cellcolor{verylightgray} .013} & \cellcolor{verylightgray} .017 & \cellcolor{verylightgray} .017 & \cellcolor{verylightgray} .015 & .060 & .025 & .092 & .024 & .020 & \textbf{.017} & .138 & .030 & .193 & .034 & .021 & \textbf{.017} \\
0.1 & \cellcolor{verylightgray} .022 & \cellcolor{verylightgray} .021 & \textbf{\cellcolor{verylightgray} .013} & \cellcolor{verylightgray} .017 & \cellcolor{verylightgray} .018 & \cellcolor{verylightgray} .015 & .061 & .025 & .094 & .024 & .020 & \textbf{.018} & .140 & .029 & .197 & .032 & .022 & \textbf{.019} \\
0.2 & \cellcolor{verylightgray} .022 & \cellcolor{verylightgray} .022 & \textbf{\cellcolor{verylightgray} .013} & \cellcolor{verylightgray} .018 & \cellcolor{verylightgray} .019 & \cellcolor{verylightgray} .016 & .063 & .025 & .096 & .023 & .021 & \textbf{.020} & .144 & .029 & .201 & .030 & .024 & \textbf{.020} \\
0.3 & \cellcolor{verylightgray} .022 & \cellcolor{verylightgray} .021 & \textbf{\cellcolor{verylightgray} .014} & \cellcolor{verylightgray} .019 & \cellcolor{verylightgray} .020 & \cellcolor{verylightgray} .018 & .067 & .024 & .098 & .022 & .024 & \textbf{.021} & .151 & .029 & .207 & .029 & .027 & \textbf{.022} \\
0.4 & \cellcolor{verylightgray} .022 & \cellcolor{verylightgray} .022 & \textbf{\cellcolor{verylightgray} .015} & \cellcolor{verylightgray} .020 & \cellcolor{verylightgray} .023 & \cellcolor{verylightgray} .020 & .072 & .025 & .102 & .023 & .028 & \textbf{.023} & .160 & .033 & .214 & .032 & .033 & \textbf{.023} \\
0.5 & \cellcolor{verylightgray} .022 & \cellcolor{verylightgray} .024 & \textbf{\cellcolor{verylightgray} .016} & \cellcolor{verylightgray} .022 & \cellcolor{verylightgray} .026 & \cellcolor{verylightgray} .023 & .079 & .028 & .107 & .026 & .034 & \textbf{.025} & .171 & .039 & .224 & .039 & .039 & \textbf{.024} \\
0.6 & \cellcolor{verylightgray} .023 & \cellcolor{verylightgray} .026 & \textbf{\cellcolor{verylightgray} .016} & \cellcolor{verylightgray} .024 & \cellcolor{verylightgray} .031 & \cellcolor{verylightgray} .028 & .085 & .033 & .112 & .031 & .040 & \textbf{.029} & .182 & .049 & .235 & .052 & .048 & \textbf{.027} \\
0.7 & \cellcolor{verylightgray} .026 & \cellcolor{verylightgray} .029 & \textbf{\cellcolor{verylightgray} .019} & \cellcolor{verylightgray} .029 & \cellcolor{verylightgray} .038 & \cellcolor{verylightgray} .035 & .090 & .040 & .118 & .039 & .051 & \textbf{.034} & .195 & .066 & .248 & .073 & .061 & \textbf{.031} \\
0.8 & \cellcolor{verylightgray} .030 & \cellcolor{verylightgray} .036 & \textbf{\cellcolor{verylightgray} .022} & \cellcolor{verylightgray} .035 & \cellcolor{verylightgray} .048 & \cellcolor{verylightgray} .046 & .098 & .053 & .121 & .053 & .067 & \textbf{.046} & .209 & .089 & .256 & .099 & .081 & \textbf{.038} \\
0.9 & \cellcolor{verylightgray} .036 & \cellcolor{verylightgray} .045 & \textbf{\cellcolor{verylightgray} .030} & \cellcolor{verylightgray} .048 & \cellcolor{verylightgray} .066 & \cellcolor{verylightgray} .065 & .104 & .070 & .127 & .068 & .088 & \textbf{.064} & .223 & .119 & .272 & .126 & .103 & \textbf{.051} \\
1.0 & \cellcolor{verylightgray} .057 & \cellcolor{verylightgray} .069 & \textbf{\cellcolor{verylightgray} .049} & \cellcolor{verylightgray} .081 & \cellcolor{verylightgray} .099 & \cellcolor{verylightgray} .116 & .090 & .090 & .118 & \textbf{.082} & .108 & .112 & .218 & .148 & .275 & .141 & .110 & \textbf{.084} \\
\bottomrule
\end{tabular}
 }}
 \caption{\label{tab:covariate025}Results for global covariate shift
 when $p_{L} = 0.25$ in terms of MAE. Each row contains the results
 for a degree in covariate shift computed as
 $(\alpha^{L}-\alpha^{U})$. Results are presented in three groups,
 depending on the prevalence used for the positive class in the test
 samples $p_{U}$. Columns with a grey background represent cases of
 global \emph{pure} covariate shift, in which $P_{L}(Y)=P_{U}(Y)$.}
\end{table}

\begin{table}[t!]
 \setlength{\tabcolsep}{2.5pt}{\resizebox{\textwidth}{!}{
 \begin{tabular}{r|rrrrrr|rrrrrr|rrrrrr}
\toprule
 & \multicolumn{6}{c}{$p_{U}=0.25$} & \multicolumn{6}{c}{$p_{U}=0.5$} & \multicolumn{6}{c}{$p_{U}=0.75$} \\
 & CC & ACC & PCC & PACC & DyS & SLD & CC & ACC & PCC & PACC & DyS & SLD & CC & ACC & PCC & PACC & DyS & SLD \\
\midrule
-1.0 & .171 & .057 & .218 & \textbf{.047} & .070 & .115 & \textbf{.054} & .085 & .075 & \ddag{.062} & .116 & .169 & \cellcolor{verylightgray} .074 & \cellcolor{verylightgray} .135 & \textbf{\cellcolor{verylightgray} .067} & \cellcolor{verylightgray} .122 & \cellcolor{verylightgray} .160 & \cellcolor{verylightgray} .194 \\
-0.9 & .168 & .046 & .218 & \textbf{.036} & .052 & .075 & .063 & .054 & .089 & \textbf{.040} & .074 & .100 & \cellcolor{verylightgray} .052 & \cellcolor{verylightgray} .083 & \textbf{\cellcolor{verylightgray} .042} & \cellcolor{verylightgray} .076 & \cellcolor{verylightgray} .097 & \cellcolor{verylightgray} .111 \\
-0.8 & .165 & .035 & .215 & \textbf{.028} & .040 & .054 & .066 & .042 & .094 & \textbf{.033} & .052 & .069 & \cellcolor{verylightgray} .041 & \cellcolor{verylightgray} .062 & \textbf{\cellcolor{verylightgray} .031} & \cellcolor{verylightgray} .054 & \cellcolor{verylightgray} .066 & \cellcolor{verylightgray} .075 \\
-0.7 & .159 & .032 & .211 & \textbf{.028} & .039 & .042 & .066 & .035 & .096 & \textbf{.029} & .046 & .050 & \cellcolor{verylightgray} .033 & \cellcolor{verylightgray} .046 & \textbf{\cellcolor{verylightgray} .023} & \cellcolor{verylightgray} .040 & \cellcolor{verylightgray} .052 & \cellcolor{verylightgray} .052 \\
-0.6 & .154 & \ddag{.030} & .207 & \textbf{.030} & .035 & .033 & .065 & .032 & .096 & \textbf{.028} & .038 & .039 & \cellcolor{verylightgray} .030 & \cellcolor{verylightgray} .038 & \textbf{\cellcolor{verylightgray} .019} & \cellcolor{verylightgray} .032 & \cellcolor{verylightgray} .040 & \cellcolor{verylightgray} .039 \\
-0.5 & .149 & .032 & .202 & .033 & .031 & \textbf{.028} & .063 & .031 & .095 & \textbf{.027} & .033 & .031 & \cellcolor{verylightgray} .028 & \cellcolor{verylightgray} .033 & \textbf{\cellcolor{verylightgray} .017} & \cellcolor{verylightgray} .028 & \cellcolor{verylightgray} .034 & \cellcolor{verylightgray} .031 \\
-0.4 & .146 & .032 & .199 & .035 & .029 & \textbf{.024} & .062 & .029 & .094 & .027 & .029 & \textbf{.026} & \cellcolor{verylightgray} .026 & \cellcolor{verylightgray} .029 & \textbf{\cellcolor{verylightgray} .015} & \cellcolor{verylightgray} .025 & \cellcolor{verylightgray} .029 & \cellcolor{verylightgray} .025 \\
-0.3 & .145 & .032 & .198 & .035 & .025 & \textbf{.022} & .064 & .029 & .094 & .026 & .025 & \textbf{.023} & \cellcolor{verylightgray} .023 & \cellcolor{verylightgray} .026 & \textbf{\cellcolor{verylightgray} .014} & \cellcolor{verylightgray} .022 & \cellcolor{verylightgray} .025 & \cellcolor{verylightgray} .021 \\
-0.2 & .144 & .032 & .195 & .036 & .023 & \textbf{.021} & .064 & .027 & .094 & .025 & .022 & \textbf{.020} & \cellcolor{verylightgray} .022 & \cellcolor{verylightgray} .024 & \textbf{\cellcolor{verylightgray} .012} & \cellcolor{verylightgray} .019 & \cellcolor{verylightgray} .021 & \cellcolor{verylightgray} .018 \\
-0.1 & .145 & .030 & .196 & .033 & .022 & \textbf{.019} & .066 & .025 & .095 & .023 & .020 & \textbf{.018} & \cellcolor{verylightgray} .020 & \cellcolor{verylightgray} .022 & \textbf{\cellcolor{verylightgray} .012} & \cellcolor{verylightgray} .018 & \cellcolor{verylightgray} .019 & \cellcolor{verylightgray} .016 \\
0.0 & .146 & .028 & .196 & .030 & .021 & \textbf{.019} & .068 & .024 & .096 & .022 & .020 & \textbf{.017} & \cellcolor{verylightgray} .019 & \cellcolor{verylightgray} .020 & \textbf{\cellcolor{verylightgray} .011} & \cellcolor{verylightgray} .016 & \cellcolor{verylightgray} .018 & \cellcolor{verylightgray} .015 \\
0.1 & .150 & .027 & .201 & .028 & .022 & \textbf{.018} & .070 & .023 & .099 & .021 & .021 & \textbf{.017} & \cellcolor{verylightgray} .019 & \cellcolor{verylightgray} .019 & \textbf{\cellcolor{verylightgray} .011} & \cellcolor{verylightgray} .016 & \cellcolor{verylightgray} .018 & \cellcolor{verylightgray} .016 \\
0.2 & .156 & .030 & .208 & .029 & .026 & \textbf{.019} & .073 & .025 & .102 & .022 & .024 & \textbf{.018} & \cellcolor{verylightgray} .019 & \cellcolor{verylightgray} .020 & \textbf{\cellcolor{verylightgray} .011} & \cellcolor{verylightgray} .017 & \cellcolor{verylightgray} .020 & \cellcolor{verylightgray} .017 \\
0.3 & .164 & .034 & .215 & .033 & .032 & \textbf{.020} & .078 & .026 & .107 & .024 & .028 & \textbf{.019} & \cellcolor{verylightgray} .020 & \cellcolor{verylightgray} .020 & \textbf{\cellcolor{verylightgray} .012} & \cellcolor{verylightgray} .017 & \cellcolor{verylightgray} .023 & \cellcolor{verylightgray} .018 \\
0.4 & .173 & .043 & .224 & .042 & .040 & \textbf{.021} & .084 & .030 & .112 & .028 & .033 & \textbf{.022} & \cellcolor{verylightgray} .020 & \cellcolor{verylightgray} .021 & \textbf{\cellcolor{verylightgray} .012} & \cellcolor{verylightgray} .018 & \cellcolor{verylightgray} .026 & \cellcolor{verylightgray} .021 \\
0.5 & .185 & .054 & .235 & .054 & .050 & \textbf{.025} & .091 & .036 & .117 & .033 & .040 & \textbf{.025} & \cellcolor{verylightgray} .021 & \cellcolor{verylightgray} .023 & \textbf{\cellcolor{verylightgray} .013} & \cellcolor{verylightgray} .019 & \cellcolor{verylightgray} .029 & \cellcolor{verylightgray} .023 \\
0.6 & .201 & .071 & .248 & .073 & .064 & \textbf{.030} & .101 & .045 & .125 & .042 & .050 & \textbf{.029} & \cellcolor{verylightgray} .023 & \cellcolor{verylightgray} .025 & \textbf{\cellcolor{verylightgray} .014} & \cellcolor{verylightgray} .021 & \cellcolor{verylightgray} .035 & \cellcolor{verylightgray} .028 \\
0.7 & .223 & .097 & .266 & .102 & .082 & \textbf{.037} & .116 & .060 & .136 & .057 & .063 & \textbf{.037} & \cellcolor{verylightgray} .026 & \cellcolor{verylightgray} .030 & \textbf{\cellcolor{verylightgray} .017} & \cellcolor{verylightgray} .024 & \cellcolor{verylightgray} .041 & \cellcolor{verylightgray} .034 \\
0.8 & .244 & .130 & .284 & .139 & .111 & \textbf{.050} & .128 & .078 & .146 & .076 & .086 & \textbf{.050} & \cellcolor{verylightgray} .030 & \cellcolor{verylightgray} .035 & \textbf{\cellcolor{verylightgray} .020} & \cellcolor{verylightgray} .029 & \cellcolor{verylightgray} .055 & \cellcolor{verylightgray} .044 \\
0.9 & .284 & .178 & .317 & .186 & .161 & \textbf{.074} & .155 & .106 & .166 & .102 & .120 & \textbf{.074} & \cellcolor{verylightgray} .037 & \cellcolor{verylightgray} .043 & \textbf{\cellcolor{verylightgray} .026} & \cellcolor{verylightgray} .035 & \cellcolor{verylightgray} .070 & \cellcolor{verylightgray} .060 \\
1.0 & .345 & .255 & .361 & .265 & .228 & \textbf{.149} & .196 & .158 & .194 & .153 & .172 & \textbf{.140} & \cellcolor{verylightgray} .059 & \cellcolor{verylightgray} .069 & \textbf{\cellcolor{verylightgray} .043} & \cellcolor{verylightgray} .059 & \cellcolor{verylightgray} .100 & \cellcolor{verylightgray} .106 \\
\bottomrule
\end{tabular}
 }}
 \caption{\label{tab:covariate075}Results for global covariate shift
 when $p_{L} = 0.75$ in terms of MAE. Each row contains the results
 for a degree in covariate shift computed as
 $(\alpha^{L}-\alpha^{U})$. Results are presented in three groups,
 depending on the prevalence used for the positive class in the test
 samples $p_{U}$. Columns with a grey background represent cases of
 global \emph{pure} covariate shift, in which $P_{L}(Y)=P_{U}(Y)$.}
\end{table}


The situation changes drastically when analysing the results for
global \textit{mixed} covariate shift (which in the tables are
represented by the columns with a white background), i.e., when also
$P(Y)$ changes across training data and test data. In these cases, the
performance of methods like PCC or CC (methods that performed very
well under the presence of global \textit{pure} covariate shift)
degrades, due to the fact that these methods do not attempt any
adjustment to the prevalence of the test data. In this case,
methods designed to deal with prior probability shift, such as SLD,
stand as the best performers. 
This is interesting, since this experiment represents a situation in
which a change in the covariates happens along with a change in the
priors, thus harming the calibration of the posterior probabilities on
which PCC rests upon.

\subsection{Local Covariate Shift}
\label{sec:exp:changes}


\begin{figure}[t!]
 \centering
 \includegraphics[width=\textwidth]{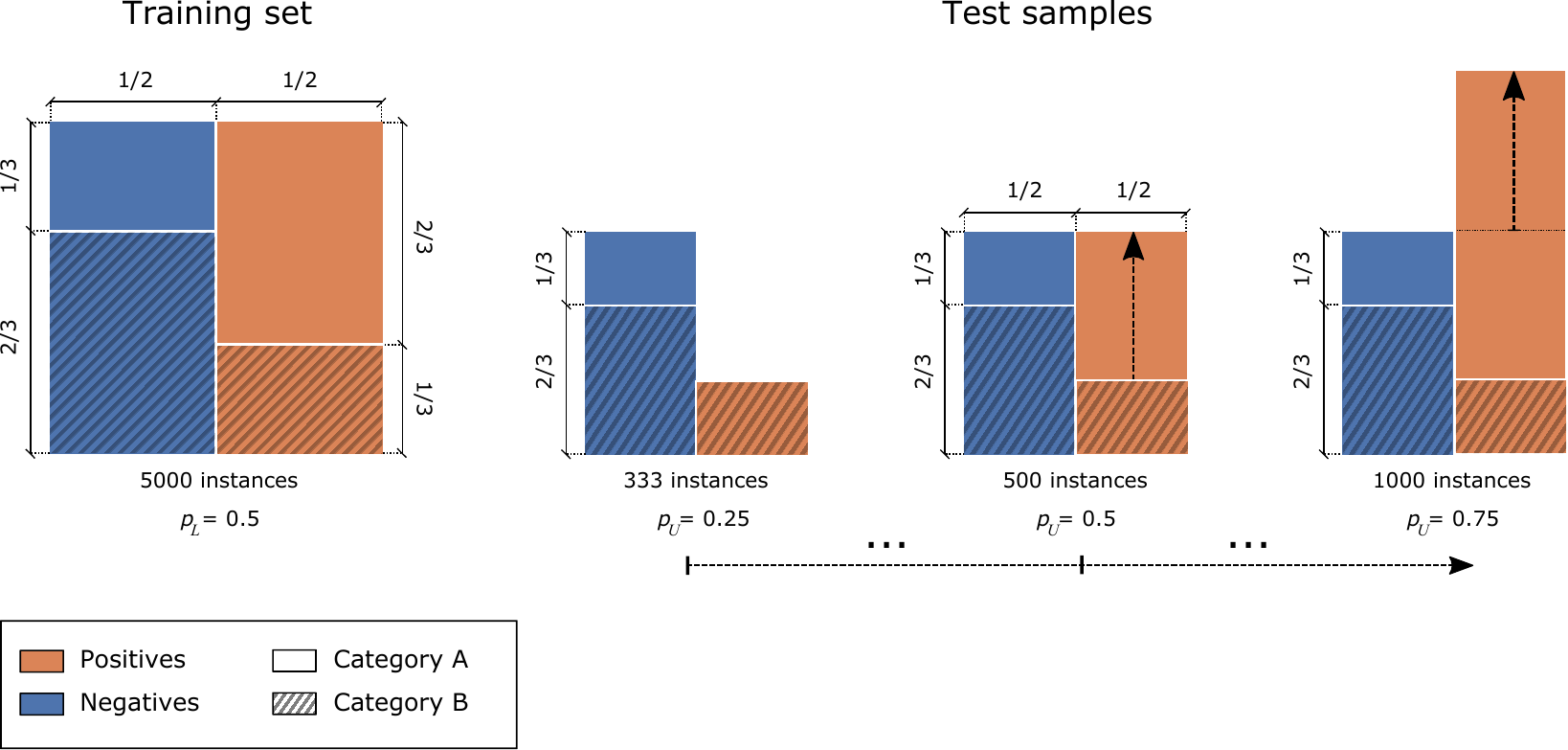}
 \caption{Conceptual diagram illustrating our local covariate shift
 protocol.}
 \label{fig:exp:localcovariatet2}
\end{figure}


\subsubsection{Evaluation Protocol}

\noindent For simulating \emph{local} covariate shift we generate a
shift in the class conditional distribution of only one of the
classes. In order to do so, categories $A$ and $B$ are treated as
subclasses, or clusters, of the positive and negative classes. Figure
\ref{fig:exp:localcovariatet2} might help in understanding this
protocol. The main idea is to alter the prevalence $P(Y)$ of the test
samples by just changing the prevalence of positive documents of one
of the subclasses (e.g., of category $A$) while maintaining the rest
(e.g., positives and negatives in $B$ and the negatives of $A$)
unchanged. Following this procedure, we let the class-conditional
distribution of the positive examples $P(X|Y=1)$ vary, while the
class-conditional distribution of the negative examples $P(X|Y=0)$
remains constant.

For this experiment, we keep the training prevalence fixed at
$p_{L}=0.5$, while we vary the test prevalence $p_{U}$ artificially.
To allow for a wider exploration of the range of the prevalence values
$p_{U}$ that can be achieved by varying only the number of positives
in category $A$, we start from a configuration in which $\frac{2}{3}$
of the positives in the training set are from category $A$ and the
remaining $\frac{1}{3}$ are from category $B$. Both categories
contribute to the training set with exactly the same number of
documents (2,500 each, since the training set contains 5,000
documents, as before). The set of negative examples is composed of
$\frac{1}{3}$ documents from $A$ and $\frac{2}{3}$ documents from $B$.
In the test samples all these proportions are kept fixed except for
the positive documents from category $A$, so that a desired prevalence
value is reached by removing, or adding, positives of this
category. Note that this process generates test samples of varying
sizes. In particular, when the test size is equal to 500, the
proportions of positive and negative documents, as well as the
proportion of documents from $A$ and $B$, match the proportions used
in the training set. Using this procedure we explore $p_{U}$ in the
range $[0.25, 0.75]$ at steps of 0.05 (see
Algorithm~\ref{alg:covariatet2}).

For this experiment the number of test samples used for evaluation
amounts to $11\times11\times50\times10$ = 60,500 for each
quantification algorithm we test.


\begin{algorithm}[t!]
 \hspace*{\algorithmicindent} \textbf{Input:} Datasets $A$ and $B$;
 Quantification algorithm $Q$ \\
 \begin{algorithmic}[1]
 \State
 $A \gets \operatorname{binarise\_dataset}(A,
 \operatorname{cut\_point=3})$ \State
 $B \gets \operatorname{binarise\_dataset}(B,
 \operatorname{cut\_point=3})$ \State
 $\mathcal{L}_A, \mathcal{U}_A \gets
 \operatorname{split\_stratified}(A)$ \State
 $\mathcal{L}_B, \mathcal{U}_B \gets
 \operatorname{split\_stratified}(B)$

 \For{10 repetitions}

 \State $L_A \sim \mathcal{L}_A$ with $p_{L_A}=\frac{2}{3}$ and
 $|L_A|=2500$ \State $L_B \sim \mathcal{L}_B$ with
 $p_{L_B}=\frac{1}{3}$ and $|L_B|=2500$ \State
 $L \gets L_A \cup L_B$ \hspace{0.5cm}{\commentfont /* Note
 $p_{L}=\frac{1}{2}$ and $|L|=5000$ */}

 \State {\commentfont /* Use quantification algorithm $Q$ to learn
 a quantifier $q$ on $L$ */}

 \State $q \gets Q.fit(L)$

 \For{50 repetitions}

 \State $U^\ominus_A \sim \mathcal{U}_A$ with $p_{U^\ominus_A}=0$
 and $|U^\ominus_A|=\frac{250}{3}$

 \State $U_B \sim \mathcal{U}_B$ with $p_{U_B}=\frac{1}{3}$ and
 $|U_A|=250$

 \State {\commentfont /* Note
 $p_{\{U^\ominus_A \cup U_B\}}=\frac{1}{4}$ */}

 \For{$p_{U} \in \{0.25, 0.3, 0.35, \ldots, 0.75\}$}

 \State solve for $\operatorname{POS}$ the equation
 $p_{U}=\frac{\frac{250}{3}+\operatorname{POS}}{\frac{250}{3}+250+\operatorname{POS}}$

 \State $U^\oplus_A \sim \mathcal{U}_A$ with $p_{U^\oplus_A}=1$ and
 $|U^\oplus_A|=\operatorname{POS}$

 \State $U \gets U^\oplus_A \cup U^\ominus_A \cup U_B$

 \State $\hat{p}_{U}^{q} \gets q.\operatorname{quantify}(U)$ \State
 $error \gets AE(p_{U},\hat{p}_{U}^{q})$

 \EndFor

 \EndFor \EndFor
 \end{algorithmic}
 \caption{The protocol for generating \emph{local} covariate shift.}
 \label{alg:covariatet2}
\end{algorithm}


\subsubsection{Results}

\begin{figure}[t!]
 \centering
 \includegraphics[width=\textwidth]{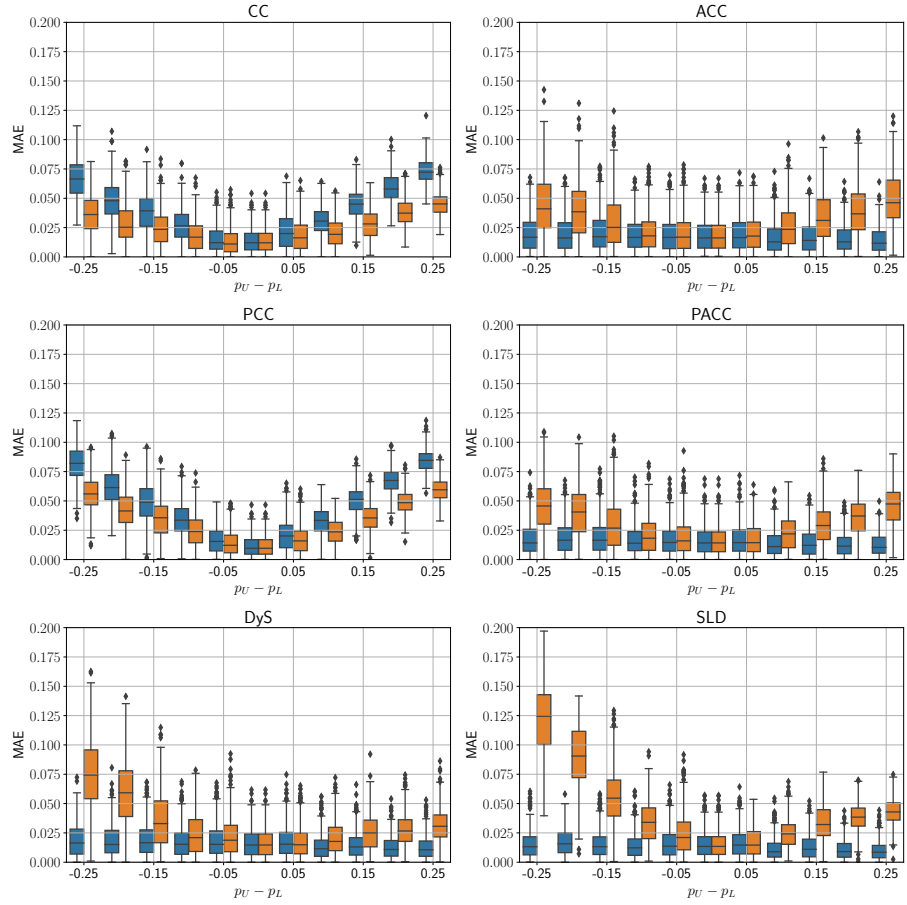}
 \caption{\label{tab:priorchangetab}Results for \emph{local}
 covariate shift expressed in terms of $\operatorname{MAE}$. Blue
 boxes represent the situation in which $P_{L}(X|Y) = P_{U}(X|Y)$
 while orange boxes represent the situation in which
 $P_{L}(X|Y) \neq P_{U}(X|Y)$ because
 $P_{L}(X|Y=1) \neq P_{U}(X|Y=1)$. The degree of shift in the priors
 is shown along the x-axis and is computed as $(p_{U}-p_{L})$ rounded
 to two decimals.}
 \label{fig:exp:covariatet2}
\end{figure}

\noindent The results we have obtained for local covariate shift
(orange boxes) are displayed in Figure~\ref{fig:exp:covariatet2}. For
easier comparison, this plot also shows results for the cases in which
the class-conditional distributions are constant across the training
data and the test data (blue boxes), i.e., when the type of shift is
prior probability shift.

Consistently with the results of Section~\ref{sec:resultsprior}, most
quantification algorithms (except for CC and PCC) work reasonably well
(see the blue boxes) when the class-conditional distributions are
invariant across the training and the test data. Instead, when the
class-conditional distributions change, the performance of these
algorithms tends to degrade. This should come at no surprise given
that all the adjustments implemented in the quantification methods we
consider (as well as in all other methods we are aware of) rely on the
assumption that the class-conditional distributions are invariant. The
exception to this are CC and PCC, the only methods that do not attempt
to adjust the priors. What comes instead as a surprise is not only
that the performance of CC and PCC does not degrade, but that this
performance seems to improve (i.e., the orange boxes in the extremes
are systematically below the blue boxes for CC and PCC). This
apparently strange behaviour can be explained as follows. When
$p_{U} \ll p_{L}$, CC and PCC will naturally tend to overestimate the
true prevalence. However, in this case, the positive examples in the
test sample happen to mostly be from category $B$. Since the
underlying classifier has been trained on a dataset in which the
positives from category $A$ were more abundant ($\frac{2}{3}$) than
the positives from category $B$ ($\frac{1}{3}$), the classifier has
more problems in classifying positives from $B$ than from category
$A$. This has the consequence that the overestimation brought about by
CC and PCC is partially compensated (that is, positive examples from
$B$ tend to be misclassified as negatives more often), and thus the
final $\hat{p}_{U}$ gets closer to the real value $p_{U}$. On the
other side, when $p_{U} \gg p_{L}$, CC and PCC will tend to
underestimate $\hat{p}$. However, in this scenario positive examples
mostly belong to category $A$, which the classifier identifies as
positives more easily (since it has been trained on a relatively
higher number of positives from $A$), thus increasing the value of
$\hat{p}_{U}$ and making it closer to the actual value $p_{U}$.

A fundamental conclusion of this experiment is that, when the
class-condi\-tio\-nal distributions change, the adjustment implemented
by the most sophisticated quantification methods can become
detrimental. This is important since, in real applications, there is
no guarantee that the type of shift a system is confronted with is
prior probability shift, nor is there any general way for reliably
identifying the type of shift involved. This experiment also shows how
the bias inherited by CC and PCC can, under some circumstances, be
``serendipitously'' mitigated, at least in part.
(We will see a similar example when studying concept shift in
Section~\ref{sec:exp:concept}.)


\subsection{Concept Shift}
\label{sec:exp:concept}


\subsubsection{Evaluation Protocol}
\label{sec:conceptshiftgeneration}

\begin{algorithm}[t!]
 \hspace*{\algorithmicindent} \textbf{Input:} Categories $A$ and $B$;
 Quantification algorithm $Q$ \\
 \begin{algorithmic}[1]
 \State {\commentfont /* Sample $\mathcal{D}$ balanced with respect
 to number of stars */} \State $D \sim A \cup B$ with
 $p_D(\mathcal{Y}_{\star})=(0.2, 0.2, 0.2, 0.2, 0.2)$ \State
 $\mathcal{L},\; \mathcal{U} \gets
 \operatorname{split\_stratified(\emph{D})}$ \For{10 repetitions}
 \For{$c^{L} \in \{1.5,2.5,3.5,4.5\}$} \State {\commentfont /*
 Generate a sample from $\mathcal{L}$ */} \State
 $L \sim \mathcal{L}$ with $|L|=5000$ \State {\commentfont /*
 Binarising using this specific cut point */} \State
 $L \gets \operatorname{binarise\_dataset}(L,
 \operatorname{cut\_point}=c^{L})$ \State {\commentfont /* Use
 quantification algorithm $Q$ to learn $q$ on $L$ */} \State
 $q \gets Q.fit(L)$ \For{50 repetitions} \State {\commentfont /*
 Generating test samples */}
 \For{$c^{U} \in \{1.5,2.5,3.5,4.5\}$} \State $U \sim \mathcal{U}$
 with $|U|=500$ \State {\commentfont /* Binarising using this
 specific cut point */} \State
 $U \gets \operatorname{binarise\_dataset}(U,
 \operatorname{cut\_point}=c^{U})$ \State
 $\hat{p}_{U}^{q} \gets q.\operatorname{quantify}(U)$ \State
 $error \gets AE(p_{U},\hat{p}_{U}^{q})$ \EndFor \EndFor \EndFor
 \EndFor
 \end{algorithmic}
\caption{Protocol for generating concept shift.
}
\label{alg:concept}
\end{algorithm}


\noindent In order to simulate concept shift we exploit the ordinal
nature of the original 5-star ratings. Specifically, we simulate
changes in the concept of ``being positive'' by varying, in a
controlled manner, the threshold above which a review is considered
positive. The protocol we propose thus comes down to varying the cut
points in the training set ($c^{L}$) and in the test set ($c^{U}$)
\emph{independently}, so that the notion of what is considered
positive differs between the two sets. For example, by imposing a
training cut point of $c^{L}=1.5$ we are mapping 1-star to the
negative class, and 2-, 3-, 4-, and 5-stars to the positive class. In
other words, everything but strongly negative reviews are considered
positive in the training set. If, at the same time, we set the test
cut point at $c^{U}=4.5$, we are generating a large shift in the
concept of ``being positive'', since in the test set only strongly
positive reviews (5 stars) will be considered positive. For 5 classes
there are 4 possible cut points $\{1.5,2.5,3.5,4.5\}$; the protocol
explores all combinations systematically (see
Algorithm~\ref{alg:concept}).

We use the signed difference $(c^{L}-c^{U})$ as an indication of the
degree of concept shift, resulting in an integer value in the range
$[-3, 3]$; note that $(c^{L}-c^{U})=0$ corresponds to a situation in
which there is no concept shift.

It is also worth noting that this protocol \emph{does not affect
$P(X)$}, which remains constant across the training distribution and
the test distribution. Conversely, varying the cut point has a direct
effect on $P(Y)$, which means that by establishing different cut
points for the training and the test datasets we are indirectly
inducing a change in the priors. In order to allow for controlled
variations in the priors, we depart from a situation in which all five
ratings have the same number of examples, i.e, we impose
$p(\mathcal{Y}_{\star})=(0.2,0.2,0.2,0.2,0.2)$ onto both the training
set and the test set. This guarantees that a change in a cut point
$c\in\{1.5,2.5,3.5,4.5\}$ gives rise to a binary set with (positive)
prevalence values in $\{0.2, 0.4, 0.6, 0.8\}$, which in turn implies a
difference in priors
$(p_{L}-p_{U})\in\{-0.6, -0.4, \ldots, 0.4, 0.6\}$.

For this experiment, the number of test samples used for evaluation
amounts to $4\times4\times50\times10$ = 8,000 for each quantification
algorithm we test.



\begin{figure}[t!]
 \begin{center}
 \includegraphics[width=\textwidth]{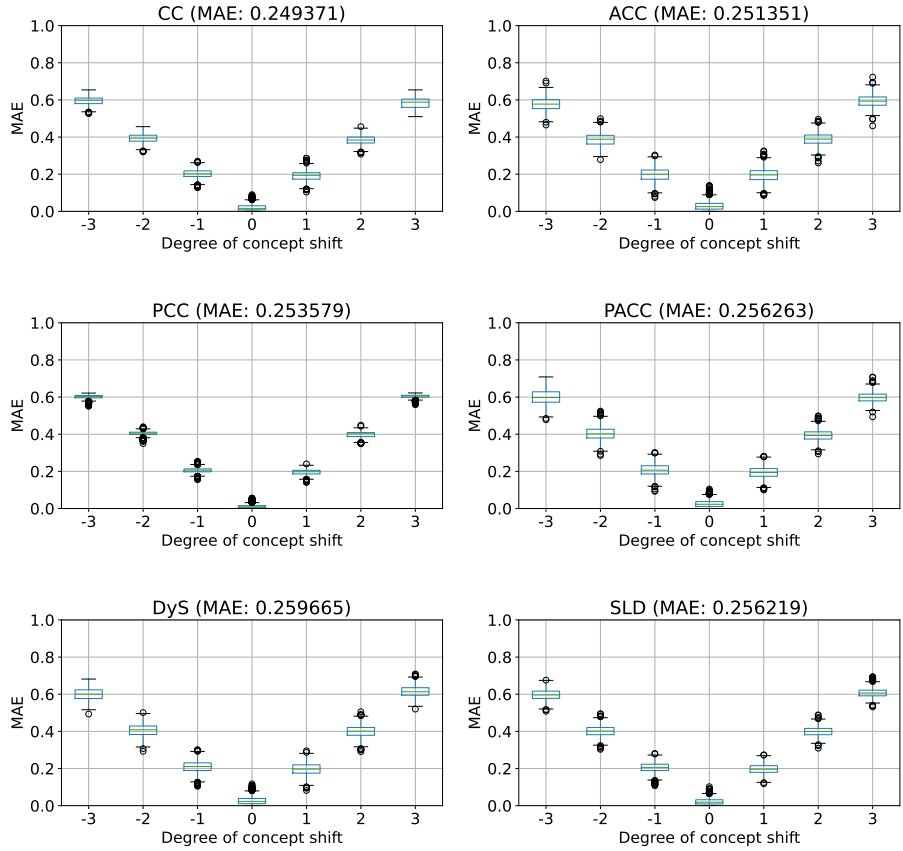}
 \end{center}
 \caption{Results for concept shift. The error measure is MAE and the
 degree of concept shift is computed as $c_{tr} - c_{tst}$.}
 \label{fig:conceptresults_v2}
\end{figure}


\subsubsection{Results}
\label{sec:exp:concept:res}


\noindent The results for our simulation of concept shift are shown in
Figure~\ref{fig:conceptresults_v2}. The performance of all methods
decreases as the degree of concept shift increases, i.e., when
$c^{L}<c^{U}$ (resp., $c^{L}>c^{U}$) all methods tend to overestimate
(resp., underestimate) the true prevalence. That no method could fare
well under concept shift was expected, for the simple reason that none
of these methods has been designed to confront arbitrary changes in
the functional relationship between covariates and classes. These
results deserve no further discussion, and are here reported only for
the sake of completeness (we omit the corresponding table, though).

\begin{figure}[t!]
 \begin{center}
 \includegraphics[width=\textwidth]{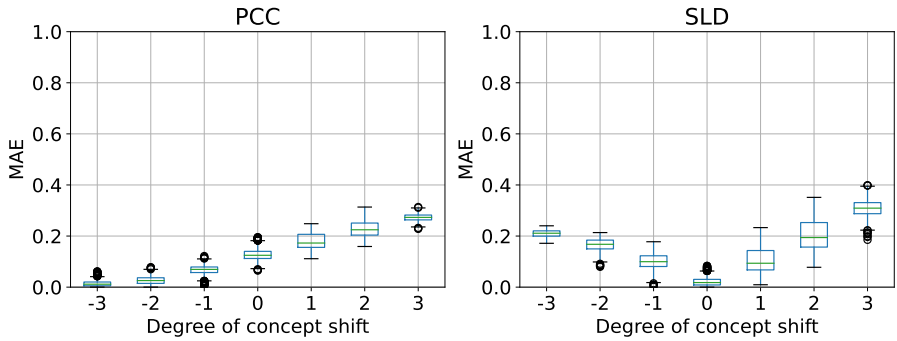}
 \end{center}
 \caption{Results for concept shift with forced values for
 $p_{L}=0.5$ and $p_{U}=0.75$. The error measure is MAE and the
 degree of concept shift is computed as $(c_{tr} - c_{tst})$.}
 \label{fig:conceptresults05}
\end{figure}

What instead deserves some discussion is the fact that concept shift
might, under certain circumstances, lead to erroneous interpretations
of the relative merits of quantification methods. This confusion might
arise when the \emph{bias} of a quantifier gets partially compensated
by the variation in the prior resulting from the change in the
concept. This situation is reproduced in
Figure~\ref{fig:conceptresults05}, where we impose $p_{L}=0.5$ and
$p_{U}=0.75$.\footnote{As a consequence of resampling, $P(X)$ changes
across the training and the test data.} Take a look at the errors
produced by both methods when $(c^{L}-c^{U})=0$, i.e., when
$c^{L}=c^{U}$. Note that in this case, there is no concept shift, but
there is prior probability shift. (Recall that we chose $p_{L}=0.5$
and $p_{U}=0.75$ for this experiment). We know that PCC tends to
deliver biased estimators, while SLD instead does not. This is
witnessed by the fact that PCC yields an error close to MAE=0.15 (it
tends to underestimate the test prevalence), while SLD obtains a very
low error instead; let us call this bias the ``global'' bias. As we
separate the cut points, we introduce a form of bias (a ``local''
bias) that interacts with the global one. For instance, imagine we
train our classifier with 1-star and 2-stars acting as negative labels
and (3, 4, 5) acting as positive ones. Assume that in test we instead
have (1, 2, 3) stars acting as the negative labels and only (4, 5) as
the positives. In this case, the classifier will now tend to classify
as positive the test examples with 3 stars. This local overestimation
will partially compensate for the global underestimation. (An
analogous reasoning applies in the other direction as well.) Note that
such an improvement is accidental, and attributing any merit to the
quantifier for this would be misleading.

\section{Conclusions}
\label{sec:conclusions}

\noindent Since the goal of quantification is estimating class
prevalence, most previous efforts in the field have focused on
assessing the performance of quantification systems in situations
characterised by a shift in class prevalence values, i.e., by prior
probability shift; in the quantification literature other types of
dataset shift have received less attention, if any. In this paper we
have proposed new evaluation protocols for simulating different types
of dataset shift in a controlled manner, and we have used them to test
the robustness to these types of shift of several representative
methods from the quantification literature. The experimental
evaluation we have carried out has brought about some interesting
findings.

The first such finding is that many quantification methods are robust
to prior probability shift but not to other types of dataset
shift. When the simplifying assumptions that characterise prior
probability shift (e.g., that the class-conditional densities remain
unaltered) are not satisfied, all the tested methods (including SLD, a
top performer under prior probability shift)
experience a marked degradation in performance.

A second observation is that, while previous theoretical studies
indicate that PCC should be the best quantification method for dealing
with covariate shift, our experiments reveal that its use should only
be recommended when the class label proportions are expected not to
change substantially (a setting that we refer to as \emph{pure}
covariate shift).

Such a setting, though, is fairly uninteresting in real-life
applications,
and our experiments show that other methods (particularly: SLD and
PACC) are preferable to PCC when covariate shift is accompanied by a
change in the priors. However, even SLD becomes unstable under certain
conditions in which both covariates and labels change. We argue that
such a setting, which we have called \emph{local} covariate shift,
shows up in many applications of interest (e.g., prevalence estimation
of plankton subspecies in sea water samples~\citep{Gonzalez:2019fh},
or seabed cover mapping~\citep{Beijbom:2015yg}, in which finer-grained
unobserved classes are grouped into coarser-grained observed classes.

Finally, our results highlight the limitations that all quantification
methods exhibit when coping with concept shift. This was to be
expected since no method can adapt to arbitrary changes in the
functional relationship between covariates and classes without the aid
of external information. The same batch of experiments also shows that
concept shift may induce a change in the priors that can partially
compensate the bias of a quantifier; however, such an improvement is
illusory and accidental, and it is difficult to envision clever ways
for taking advantage of this phenomenon.

Possible directions for future work include extending the protocols we
have devised to other specific types of shift that may be
application-dependent (e.g., shifts due to transductive active
learning~\citep{Kottke:2022oy}, to oversampling of positive training
examples in imbalanced data scenarios~\citep{Moreo:2016hj}, to concept
shifts in cross-lingual applications), and to types of quantification
other than binary (e.g., multiclass, ordinal, multi-label). The goal
of such research, as well of the research presented in this paper, is
to allow a correct evaluation of the potential of different
quantification methods when confronted with the different ways in
which the unlabelled data we want to quantify on differs from the
training data, and to stimulate research in new quantification methods
capable of tackling the types of shift that current methods are
insufficiently equipped for.










\section*{Acknowledgments}

\noindent The work of the 1st author hast been funded by MINECO
(Ministerio de Econom\'ia y Competitividad) and FEDER (Fondo Europeo
de Desarrollo Regional), grant PID2019-110742RB-I00 (MINECO/FEDER),
and by Campus de Excelencia Internacional in collaboration with
Santander Bank in the framework of the financial aid for mobility of
excellence for teachers and researchers at the University of
Oviedo. The work by the 2nd and 3rd authors has been supported by the
\textsf{SoBigData++} project, funded by the European Commission (Grant
871042) under the H2020 Programme INFRAIA-2019-1, by the
\textsf{AI4Media} project, funded by the European Commission (Grant
951911) under the H2020 Programme ICT-48-2020, and by the
\textsc{SoBigData.it}, \textsc{FAIR}, and \textsc{QuaDaSh} projects 
funded by the Italian
Ministry of University and Research under the NextGenerationEU
program. The authors' opinions do not necessarily reflect those of the
funding bodies.



\bibliography{Fabrizio} \bibliographystyle{spbasic}

\mbox{} \label{pageref:endofpaper}


\clearpage
\appendix

\section{The Equivalence of SMM and PACC}
\label{sec:smmpacceq}

\noindent In this section we prove that the method Sample Mean
Matching (SMM) proposed by \cite{Hassan:2020kq} is equivalent to the
method Probabilistic Adjusted Classify \& Count (PACC) presented by
\cite{Bella:2010kx}. This equivalence between the two methods was
already hinted at in~\citep{Castano:2023mb} but no formal proof was
provided.

SMM fits in the DyS framework of \cite{Maletzke:2019qd},
replacing histograms, binning the posterior probabilities issued by a
soft classifier $s$, with the mean of these posteriors, and adopting
$L_{1}$ as the dissimilarity function $DS$:
\begin{align}
 \hat{p}_{\sigma}^{\operatorname{SMM}} = \argmin_{0 \leq \alpha 
 \leq 1}{|(\alpha \EX_{\mathbf{x}\in L^{\oplus}}\left[s(\mathbf{x})\right]
 +(1-\alpha)\EX_{\mathbf{x}\in L^{\ominus}}\left[s(\mathbf{x})\right])-\EX_{\mathbf{x}\in \sigma}\left[s(\mathbf{x})\right]|}
\end{align}
%

\noindent Solving for $\alpha$ when the $L_{1}$ distance is equal to 0
we obtain
\begin{align}
 \label{eq:smmproof}
 \hat{p}_{\sigma}^{\operatorname{SMM}}=\frac{\EX_{\mathbf{x}\in \sigma}\left[s(\mathbf{x})\right]-\EX_{\mathbf{x}\in L^{\ominus}}\left[s(\mathbf{x})\right]}{\EX_{\mathbf{x}\in L^{\oplus}}\left[s(\mathbf{x})\right]-\EX_{\mathbf{x}\in L^{\ominus}}\left[s(\mathbf{x})\right]}.
\end{align}
\noindent On the other hand PACC solves the following equation to
compute $\hat{p}^{\operatorname{PACC}}$:
\begin{align}
 \label{eq:paccproof}
 \hat{p}_{\sigma}^{\operatorname{PACC}} = \frac{\hat{p}_{\sigma}^{\operatorname{PCC}}-\hat{\operatorname{fpr}}_s}{\hat{\operatorname{tpr}}_s-\hat{\operatorname{fpr}}_s}
\end{align}
\noindent Both Equations~\ref{eq:smmproof} and \ref{eq:paccproof} are
equal, as all their terms are equivalent:
\begin{align}
 \EX_{\mathbf{x}\in \sigma}\left[s(\mathbf{x})\right] = & \ \frac{1}{|\sigma|}\sum_{x \in \sigma}{s(\mathbf{x})} \equiv \hat{p}_{\sigma}^{\operatorname{PCC}} \\
 \EX_{\mathbf{x}\in L^{\ominus}}\left[s(\mathbf{x})\right] = & \ \frac{1}{|L^{\ominus}|}\sum_{x \in L^{\ominus}}{s(\mathbf{x})} \equiv \hat{\operatorname{fpr}}_s \\
 \EX_{\mathbf{x}\in L^{\oplus}}\left[s(\mathbf{x})\right] = & \ \frac{1}{|L^{\oplus}|}\sum_{x \in L^{\oplus}}{s(\mathbf{x})} \equiv \hat{\operatorname{tpr}}_s
\end{align}

\end{document}